\documentclass[mnsc]{informs3} 

\usepackage[commandnameprefix=always,markup=underlined]{changes}

\DoubleSpacedXII

\usepackage{natbib}
 \bibpunct[, ]{(}{)}{,}{a}{}{,}
 \def\bibfont{\small}%
\usepackage{caption}
\usepackage{fvextra}            

\usepackage{float}
\usepackage{csquotes}
\usepackage{xcolor}
\usepackage{soul}
\usepackage[markup=underlined]{changes}
\PassOptionsToPackage{hyphens}{url}
\usepackage[colorlinks=true,linkcolor=blue,citecolor=blue]{hyperref}
\hypersetup{
    colorlinks=true,
    linkcolor=blue,
    filecolor=magenta,      
    urlcolor=cyan,
}
\usepackage[most]{tcolorbox}

\tcbset {
  base/.style={
    arc=0mm, 
    bottomtitle=0.5mm,
    boxrule=0mm,
    colbacktitle=black!10!white, 
    coltitle=black, 
    fonttitle=\bfseries, 
    left=2.5mm,
    leftrule=1mm,
    right=3.5mm,
    title={#1},
    toptitle=0.75mm, 
  }
}
\usepackage{xcolor}
\usepackage{soul}

\usepackage{xifthen}

\definecolor{brandblue}{rgb}{0.34, 0.7, 1}
\newtcolorbox{mainbox}[1]{
  colframe=lightgray, 
  base={#1}
}

\newtcolorbox{subbox}[1]{
  colframe=black!30!white,
  base={#1}
}
\tcbuselibrary{listings}

\newtcblisting{mainboxverb}{
  colframe=lightgray,
  colback=white,
  listing only,
  breakable,
  listing options={
    basicstyle=\ttfamily\small,
    breaklines=true
  }
}
\usepackage{xcolor}
\usepackage{fancyvrb}
\DefineVerbatimEnvironment{shadedverbatim}{Verbatim}{
  breaklines=true,
  fontsize=\small,
  bgcolor=gray!12
}
\TheoremsNumberedThrough     
\ECRepeatTheorems

\EquationsNumberedThrough    

\usepackage{verbatim}
\usepackage{multirow}
\usepackage{eurosym}

\hypersetup{
    colorlinks=true,
    linkcolor=blue,
    filecolor=magenta,      
    urlcolor=cyan,
}
\usepackage{array}
\usepackage[flushleft]{threeparttable}
\usepackage{enumerate}
\usepackage[font=footnotesize,labelfont=bf]{caption}
\usepackage{makecell}
\usepackage{amsmath}
\usepackage{amsfonts}
\usepackage{graphicx}
\usepackage{subcaption}
\usepackage{siunitx}
\usepackage{booktabs}
\usepackage{placeins}
\usepackage{float,morefloats}
\usepackage{tabularx}
\usepackage{tikz}
\usepackage{array,multirow}
\usepackage{booktabs}
\newcolumntype{L}[1]{>{\raggedright\let\newline\\\arraybackslash\hspace{0pt}}m{#1}}
\newcolumntype{C}[1]{>{\centering\let\newline\\\arraybackslash\hspace{0pt}}m{#1}}
\newcolumntype{R}[1]{>{\raggedleft\let\newline\\\arraybackslash\hspace{0pt}}m{#1}}
\usepackage{color}
\usepackage[capitalise]{cleveref}

\newcolumntype{K}[1]{>{\raggedright\arraybackslash}p{#1}}
\renewcommand{\theARTICLETOP}{}

\newcommand{\Methodshort}{GER}
\RRHFirstLine{ }
\LRHFirstLine{ }
\ECRRHFirstLine{ }
\ECLRHFirstLine{ }

\RRHSecondLine{\textit{Can Advanced LLMs Coach Smaller LLMs? }}

\LRHSecondLine{\textit{Can Advanced LLMs Coach Smaller LLMs? }}

\ECRRHSecondLine{\textit{Can Advanced LLMs Coach Smaller LLMs? }}

\ECLRHSecondLine{\textit{Can Advanced LLMs Coach Smaller LLMs? }}



\makeatletter
\providecommand{\@openbib@code}{}%
\makeatother

\begin{document} 




\RUNAUTHOR{Wang et~al.} %

\RUNTITLE{Library-based Interpretable Knowledge Distillation}

 \TITLE{\Large Using Better LLMs to Teach Lesser LLMs:  \\``Strategy'' Knowledge Distillation for Dynamic  Prompting}
 \TITLE{\Large Using Better LLMs to Teach Lesser LLMs:  \\A ``Strategy'' Knowledge Distillation Approach}
\TITLE{\Large Interpretable ``Strategy'' Knowledge Distillation \\to Teach Lesser LLMs Using Better LLMs}

\TITLE{\Large Using Better LLMs to Teach Lesser LLMs:\\ Interpretable Knowledge Distillation}

\TITLE{\Large Using Better LLMs to Teach Lesser LLMs:\\ Interpretable Knowledge Distillation via Strategy Teaching}

\TITLE{\Large Distilling Knowledge from Advanced LLMs: \\An Interpretable Approach to Enhance Smaller LLMs}

\TITLE{\Large Using Advanced LLMs to Enhance Smaller LLMs: \\An Interpretable Knowledge Distillation Approach }

\TITLE{\Large Knowledge Distillation through Contrastive Reasoning: \\ An Application to Conversational AI for Customer Service}

\TITLE{\Large Knowledge Distillation through Reasoning Imitation: \\ An Application to Conversational AI for Customer Service}

\TITLE{\Large Using Advanced LLMs to Enhance Smaller LLMs: \\Interpretable Knowledge Distillation for Customer Service}

\TITLE{\Large Can Stronger LLMs Coach Weaker LLMs? \\Knowledge Distillation for Goal-Oriented Dialogs}

\TITLE{\Large Coaching AI with Strategic Textual Guidance: \\A Framework for Knowledge Distillation}

\TITLE{\Large Coaching AI: Knowledge Distillation\\via Inference Time Guidance}

\TITLE{\Large From Output Imitation to Strategic Guidance: \\Knowledge Distillation for
Interactive AI Agents}

\TITLE{\Large Beyond Mimicry to Contextual Guidance: \\Knowledge Distillation for
Interactive AI}



\ARTICLEAUTHORS{%
\AUTHOR{Tong Wang and K. Sudhir
}
\AFF{Yale School of Management\\ \EMAIL{tong.wang.tw687@yale.edu, k.sudhir@yale.edu}
}
}

\ABSTRACT{

As large language models increasingly mediate firm–customer interactions, firms face a tradeoff: the most capable models perform well but are costly and difficult to control at scale. Existing knowledge distillation methods address this challenge by training weaker, deployable models to imitate frontier outputs; however, such open-loop approaches are poorly suited to interactive, multi-turn settings where responses must be sequenced coherently across conversational states. 
We propose a shift in what knowledge is distilled—from output imitation to contextual guidance. We develop a framework in which a superior teacher model constructs a reusable library of strategic textual guidance for particular scenarios likely to be encountered by the student. When deployed, the student retrieves the context-specific guidance at inference time, enabling adaptive behavior without retraining.
Using customer-service interactions, we show that this approach improves service quality and customer satisfaction relative to standard fine-tuning while maintaining alignment with firm policies. The results position inference-time textual guidance as a scalable and controllable approach to distillation for interactive AI agents in marketing settings.

}


\maketitle
\thispagestyle{empty}

%

\newpage
\pagenumbering{arabic}

\section{Introduction}\label{sec:intro}

Advances in large language models (LLMs) have rapidly expanded the scope of language-based marketing applications, enabling systems that generate content, support customers, negotiate, and assist decision-making. Yet while frontier LLMs deliver remarkable performance, deploying them at scale poses practical challenges. These models are expensive to operate,  raise privacy and reliability concerns, and limit operational control. As a result, for many production tasks, organizations prefer smaller, cheaper, and more controllable models that are easier to deploy and govern. 

This tradeoff between performance and deployability creates a recurring learning problem: how can operationally preferred but weaker models acquire the competencies of stronger ones? More broadly, how can firms transfer the capabilities of frontier models into systems that are practical to deploy without sacrificing performance? This challenge is especially acute in interactive applications such as selling, service, customer support, and negotiations. Unlike one-shot generation tasks, these settings require sequential decision-making that anticipates and adapts to an evolving conversation. They therefore benefit most from frontier-level capabilities, yet their high token volumes and privacy and control constraints make reliance on frontier models often impractical. For the firm, this paradox creates a strategic risk: the pursuit of operational efficiency via smaller models often comes at the expense of service quality and brand consistency.

Addressing this tension raises a design question: how should model behavior be adapted and controlled in practice? Two broad design paradigms have emerged for improving LLM behavior. The first relies on \emph{parameter-based adaptation}, such as supervised fine-tuning or preference-based alignment, which updates model weights using high-quality examples or annotations. This approach has become the default mechanism for steering model behavior in marketing and other applied settings \citep{toubia2505twin,timoshenko2025can,reisenbichler2026applying}. The second relies on \emph{inference-time, text-based control}, where prompts or contextual instructions guide behavior without modifying parameters. Growing evidence suggests that carefully structured textual guidance can also serve as a systematic mechanism for shaping behavior \citep{snell2024scaling}.


Although parameter-based learning is effective for many one-shot tasks, it is less well suited to interactive, sequential decision-making environments. In these settings, the critical judgments and tactics that drive performance—such as when to probe, reassure, escalate, or reframe—are typically \textit{implicit} rather than explicitly labeled in the data. Fine-tuning treats responses as endpoints to be imitated, leaving the decisive tactical principles embedded only indirectly in model parameters. As a result, models may reproduce observable patterns of behavior without explicitly representing which actions are strategically consequential. When interactions unfold in ways that differ from prior examples, this implicitness limits adaptation and robustness. 

More fundamentally, parameter updates operate as an \emph{open-loop} learning mechanism: models are trained offline on historical data and then deployed with fixed behavior. Multi-turn interactions, however, are path-dependent: early actions shape subsequent states and small errors can compound over time. Such environments therefore benefit from \emph{closed-loop, state-contingent feedback} that adapts decisions as the interaction unfolds. Inference-time adaptation—where behavior is conditioned on the current conversational state rather than solely on past data—provides a mechanism for implementing such closed-loop control and is therefore conceptually better aligned with the structure of the problem.

Parameter-based approaches also embed learned behaviors directly in model weights, making them difficult to inspect, revise, or selectively update after deployment. In enterprise applications—where brand identity is paramount, policies evolve, and compliance requirements change—this lack of transparency is a significant barrier to AI adoption.


\medskip
\noindent\textbf{Learning from Stronger Models: Knowledge Distillation}

When stronger models exhibit the desired expertise, a natural strategy is to transfer that expertise to weaker but deployable ones. Knowledge distillation provides a general framework for such transfer, enabling a weaker “student” to learn from a stronger “teacher” \citep{hinton2015distilling, xu2024survey}. Distillation clarifies \emph{who} learns from whom, but remains agnostic about both \emph{what} knowledge is transferred and \emph{how} it is represented.

In contemporary LLM practice, this transfer is typically implemented by training students to reproduce teacher outputs—for example, through fine-tuning on generated responses, demonstrations, or reasoning traces. Such approaches emphasize imitation of surface behavior rather than explicit representation of the underlying judgment, tactics or decision principles that govern effective interactions. Moreover, the learned behavior is encoded implicitly in model parameters. Consequently, even distillation largely relies on parameter-based mechanisms and inherits their limitations in sequential, state-contingent environments, where transparency, modularity, and adaptability at deployment are critical.

\medskip
\noindent\textbf{Guidance-Based Distillation via Textual Feedback}

We introduce an inference-time distillation mechanism designed for interactive decision problems. Rather than embedding knowledge implicitly in model weights, we represent teacher expertise explicitly as \emph{state-contingent textual guidance} that conditions the student’s behavior during interaction.

Our approach builds on two ideas. First, because distillation does not prescribe how knowledge must be represented, teacher expertise can be transferred in richer forms than output imitation alone. We therefore distill the teacher’s contextual judgment and decision logic directly, rather than only its responses. Second, recent work on prompt optimization shows that natural-language feedback can serve as a principled mechanism for improving behavior without modifying parameters. Textual feedbacks act as structured improvement signals—sometimes described as ``textual gradients''—that iteratively refine model behavior \citep{li2025survey, yuksekgonul2025optimizing, agrawal2025gepa}. We use this mechanism during training to elicit and refine high-quality guidance from the teacher, and at deployment the student simply retrieves and applies this guidance without further optimization.

By externalizing knowledge into text and applying it at inference time, our approach enables transparent inspection, modular updates, and state-contingent control in sequential and interactive  environments.

\noindent\textbf{The GER Framework}

Building on the preceding design principles, we propose \textsc{GER} (Guidance Elicitation and Retrieval), an inference-time distillation framework that transfers teacher expertise through \emph{explicit contextual textual guidance}: concise, context-specific instructions that specify how the student should act in a given conversational state.

\textbf{Guidance elicitation.}
Guidance is learned through an iterative teacher–student procedure inspired by prompt optimization. For each dialog scenario, the student first generates a response, after which the teacher compares it to its own behavior and produces targeted natural-language feedback in the form of guidance specifying how the student should adjust its reasoning. The guidance is refined over successive rounds until performance stabilizes. In this process, textual feedback acts as structured improvement signals— “textual gradients’’—that progressively sharpen the guidance. The resulting instructions encode the teacher’s decision logic in an explicit, reusable form and are stored externally as a library of $(\text{state}, \text{guidance})$ pairs.

\textbf{Guidance retrieval.}
At inference time,  given a new interaction, the system retrieves the most relevant guidance based on semantic similarity to the current conversational state from the library and conditions the student’s response accordingly. This retrieval mechanism provides state-contingent, closed-loop control: guidance is applied only when relevant and can adapt dynamically as the interaction unfolds.

Because the guidance remains external to model parameters, it can be inspected, edited, or updated by human managers without retraining, yielding a modular and transparent architecture, well-suited for brand deployment.

\textbf{Coverage of the state space.}
A practical requirement is that the guidance library cover the states the student may encounter at deployment. In sequential settings, this coverage is non-trivial: while students are guided to mimic the teacher, they never do it perfectly; thus, small deviations in early actions can compound over time, steering the interaction into regions of the state space rarely observed in expert demonstrations. As a result, guidance learned solely from static teacher trajectories (or observed real-world conversations) may fail precisely in the off-distribution states induced by the student’s own behavior. This failure mode mirrors the well-known limitations of behavioral cloning \citep{pomerleau1991efficient}, where imitation from expert demonstrations leaves the learner unequipped to recover once it drifts away from the expert’s induced state distribution.

We therefore adopt a closed-loop scenario construction strategy inspired by interactive imitation learning \citep{ross2011reduction}. Training begins with teacher-led interactions to anchor high-quality guidance and progressively incorporates student-generated trajectories, expanding coverage to the learner’s evolving state space. By iteratively collecting guidance on the states the student actually visits, the library remains aligned with deployment conditions and supports robust inference-time control.

In summary, \textsc{GER} learns strategic guidance from a teacher through iterative textual feedback, stores this guidance externally, and retrieves it at inference time to provide state-contingent control during interaction.

\medskip
\noindent\textbf{Application and Contributions}

We evaluate \textsc{GER} in the context of goal-oriented, multi-turn customer service dialogs
\citep{bordes2017learning}, a canonical example of interactive decision-making where early actions
shape subsequent states and outcomes depend critically on strategic sequencing, emotional
management, and policy adherence. This setting provides a demanding and practically relevant
testbed for assessing whether inference-time guidance can transfer expert judgment to weaker models.

Our empirical analysis addresses three questions.
First, does inference-time, scenario-specific strategic guidance improve performance relative to parameter-based approaches such as supervised fine-tuning or generic prompting?
Second, do these gains reflect improvements in the underlying drivers of customer satisfaction—such as reliability, empathy, and responsiveness—rather than merely higher average ratings?
Third, is guidance-based distillation robust to deployment realities, including distribution shift, synthetic training data, heterogeneous student architectures, and transfer across interaction contexts without retraining?

Across human evaluations and LLM-judged metrics, we find consistent evidence that guidance-based
distillation substantially improves weaker models, often matching or exceeding fine-tuning while
remaining interpretable and controllable. Improvements arise across multiple service-quality
dimensions, indicating that guidance corrects high-impact tactical decisions rather than superficial
style differences. Notably, gains persist for reasoning-capable models, suggesting that retrieved
guidance complements internal reasoning while avoiding the latency of test-time reflection.
Finally, staged scenario curation proves critical for robustness: involving the student during
synthetic dialog generation improves coverage of deployment states and yields stronger performance
than teacher-only or static data.

This paper makes three contributions. 
First, we introduce contextual-guidance-based distillation, reframing adaptation as the transfer of
contextual textual guidance applied at inference time rather than embedding knowledge in parameters.
Second, we provide empirical evidence that such guidance improves multi-turn service performance
across multiple quality dimensions and diverse student models, including reasoning-capable systems.
Third, we show that effective deployment requires coverage-driven scenario curation, and develop a
staged teacher–student generation strategy that mitigates distribution shift and enables reliable
learning even when training relies largely on synthetic data.

The remainder of the paper proceeds as follows. Section~\ref{sec:literature} reviews related work.
Section~\ref{sec:framework} presents \textsc{GER}. 
Section~\ref{sec:experiment} reports the empirical evaluation.
Section~\ref{sec:conclusion} concludes.

\section{Related Literature}\label{sec:literature}

Our work lies at the intersection of emerging applications of large language models (LLMs) in marketing and methods for transferring capabilities from strong proprietary models to weaker, more deployable models in multi-turn, goal-oriented interactions. Because customer service is inherently interactive and evaluative, we emphasize approaches that transfer \emph{strategic judgment}---how to evaluate a dialog state and prioritize tradeoffs---rather than treating distillation as the imitation of isolated outputs.

\subsection{LLMs in Marketing and Customer Service}

Recent research has begun to explore the use of LLMs across a wide range of marketing tasks, including market research and preference elicitation \citep{li2024frontiers, brand2023using, gui2023challenge}, customer needs identification \citep{timoshenko2025can}, content generation and optimization \citep{wang2025blessing, angelopoulos2024causal}, experimentation and A/B testing \citep{ye2025lola}, and recommendation and personalization systems \citep{wang2025blessing}. This growing literature highlights both the promise of LLMs as flexible tools for marketing analysis and automation and the challenges associated with aligning model behavior with managerial objectives and institutional constraints \citep{gui2023challenge}.

Relative to many of these applications, interactive applications are particularly demanding on the AI systems due to their multi-turn nature that creates path-dependence: early responses shape customer emotions, subsequent disclosures, and perceptions of fairness, and errors can be difficult to recover from in later turns. Moreover, firms must jointly achieve customer satisfaction with policy compliance, cost control, and brand consistency in real time. In contrast to  one-shot prediction or content-generation tasks, the expert evaluates a situation and anticipates downstream consequences---rather than merely imitating expert outputs. Hence the need to transfer context specific \emph{strategic guidance} as knowledge. 


\subsection{Knowledge Distillation and Guidance-Based Transfer}

Knowledge distillation (KD) originated as a framework for transferring capabilities from a large, complex model to a smaller, more efficient one \citep{hinton2015distilling}. In the context of LLMs, KD has expanded beyond soft-target matching to include transferring knowledge through diverse supervision signals such as preferences or rankings \citep{cui2024ultrafeedback} and intermediate representations or features \citep{kim2023token}. Prompts are often used to elicit task-relevant behavior from proprietary LLMs, which is then transferred to student models through various training pipelines \citep{ding2022gpt, he2024annollm}.

Much of this literature evaluates success via improvements on benchmark tasks, where performance can be assessed independently at each input. In contrast, in sequential and evaluative settings such as customer-service dialogs, errors arise less from local misprediction than from incorrect judgment over evolving dialog states---errors that compound over time and can irreversibly degrade outcomes. From this perspective, the appropriate object of distillation is not simply a mapping from inputs to outputs, but expert judgment about how to evaluate a situation, prioritize tradeoffs, and anticipate downstream consequences.

A related coaching intuition appears in feedback-based KD methods, where the teacher evaluates or ranks student outputs and uses this feedback to guide improvement, for example through reinforcement learning from AI feedback \citep{bai2022constitutional} or curated preference datasets \citep{cui2024ultrafeedback}. While effective in many settings, these approaches typically convert feedback into parameter updates, embedding transferred knowledge into model weights. Our approach differs in treating feedback as a source of explicit, reusable \emph{textual guidance}. Rather than prescribing what response to produce, guidance encodes how an expert assesses a dialog state and what strategic considerations should dominate at that point in the interaction, enabling interpretability, auditability, and rapid updating without retraining.

\subsection{Prompt Optimization and Textual Gradient Methods}

Our guidance elicitation procedure is related to work on prompt optimization, which treats prompts as editable text objects and improves model behavior through iterative refinement using natural-language feedback rather than numerical gradients. In this paradigm, learning occurs through structured modification of textual inputs that condition model outputs, rather than through updates to model parameters \citep{pryzant2023automatic, chang2023learning, yuksekgonul2025optimizing, agrawal2025gepa}. Prompt optimization thus constitutes a systematic mechanism for improving model behavior that is distinct from, but conceptually parallel to, parameter-based learning.

We build on this insight but differ in objective and scope. Prior prompt-optimization methods typically aim to improve task performance for a specific prompt or instance and may rely on online refinement or self-reflection loops that add inference-time computation. In contrast, we use natural-language guidance to refine \emph{scenario-specific guidance} offline, producing reusable scenario-specific guidance that encode expert judgment and can be retrieved when similar situations arise. This emphasis on offline refinement and reuse aligns with the latency and scalability constraints of high-volume customer service and reframes textual gradients as a mechanism for \emph{knowledge distillation} rather than one-shot prompt tuning.

\subsection{Multi-Turn Dialogs and Student-in-the-Loop Curation}

Goal-oriented dialog systems are designed to help users accomplish specific tasks through structured, multi-turn interactions \citep{bordes2017learning}. While LLMs have enabled more flexible end-to-end dialog systems \citep{snell2022context, li2023metaagents}, multi-turn interactions remain challenging, particularly for weaker models \citep{cheng2024cooper, deng2023prompting}. Small mistakes early in a conversation can shift the dialog into states that differ substantially from those seen during training, leading to compounding error and degraded performance; consistent with this challenge, recent work finds that fine-tuning can be ineffective in improving multi-turn abilities \citep{bai2024mt}.

Prior work addresses these challenges through strategy supervision, online expert consultation or planning at inference time \citep{zhang2023ask, yu2023prompt}, or fine-tuning on synthetic dialogs \citep{ulmer2024bootstrapping}. While effective in some contexts, these approaches can increase inference-time cost or rely heavily on output imitation, limiting robustness to student-induced distribution shift. Our approach instead emphasizes offline distillation of expert guidance and staged scenario generation that involves both teacher and student models. By allowing the student to generate dialogs while relying on the teacher to provide guidance, this student-in-the-loop curation exposes the expert to precisely those states where the student’s judgment is most likely to fail, ensuring that guidance is learned and retrieved when it is most consequential for multi-turn performance.

\section{The Guidance Elicitation and Retrieval Framework}\label{sec:framework}

This section presents \emph{Guidance Elicitation and Retrieval (GER)}, a guidance-based 
knowledge distillation framework for transferring strategic expertise from a strong
teacher model to a weaker, more deployable student.
Rather than improving the student through parameter updates, GER learns reusable,
interpretable \emph{textual guidance artifacts} that are applied selectively at inference
time.

As discussed in the introduction, the core observation motivating GER is that, in multi-turn service interactions,
performance depends less on reproducing average responses and more on making a small
number of high-impact tactical decisions correctly in context—for example, when to
acknowledge emotion, probe for information, escalate appropriately, or sequence remedies.
Crucially, these strategic judgments are typically \emph{implicit} in the observable data. Training corpora contain final responses, but not explicit annotations of the underlying intent, tactical reasoning, or state-contingent principles that generated them. As a result, parameter-based imitation primarily captures surface realizations of behavior rather than the latent strategic structure that drives successful outcomes.
GER therefore distills a teacher model’s expertise into an external library of
\emph{scenario–guidance} pairs that capture this strategic knowledge and can be retrieved
whenever similar situations arise.

\subsection{Overview}

Let $y_\mathcal{S}$ denote a student language model and $y_\mathcal{T}$ denote a stronger
teacher model.
GER constructs an external library
$\mathcal{L}=\{(s_i, g_i)\}_{i=1}^N$
of \emph{scenario–guidance} pairs.

A \emph{scenario} $s$ consists of an ongoing customer–agent dialog terminating with the
customer’s most recent utterance.
The associated guidance $g$ specifies \emph{how} the student should respond in that
situation—focusing on strategy and tactics rather than providing an output to imitate.
Examples include directives such as
``acknowledge the customer’s frustration before proposing solutions'' or
``offer remedies in increasing order of cost.''

GER operates in two distinct phrases: 

(i) \textit{Training (Guidance Elicitation).}
The teacher evaluates the student’s response relative to its own, generating targeted natural-language feedback that explicates the underlying decision logic and strategic intent required for behavioral alignment.
This feedback is iteratively refined until it reliably improves performance, yielding
persistent, scenario-level guidance. 

(ii) \textit{Inference (Retrieval-Based Transfer).}
At deployment, the student retrieves guidance from $\mathcal{L}$ based on similarity between
the current interaction and stored scenarios, and incorporates this guidance into its
prompt.
Behavior is therefore shaped contextually at inference time, without modifying model
parameters.

The output of GER is thus not a fine-tuned model, but an interpretable and auditable
library of reusable strategic knowledge that can be used by the original student model contextually at inference.

\subsection{Guidance Elicitation as Textual Optimization}\label{sec:ger_math}

Conceptually, GER treats guidance elicitation as an optimization problem over a
\emph{textual object}.
Unlike traditional distillation, which adjusts billions of numerical parameters, we directly refine a concise natural-language representation that encodes strategic advice. To optimize this object, we require a feedback signal that is itself compatible with natural language.

This perspective aligns with recent work showing that natural-language feedback can act as
directional supervision signals—sometimes described as \emph{textual gradients}—that
systematically improve system behavior even in the absence of differentiability
\citep{yuksekgonul2025optimizing}.
Rather than calculating a vector in parameter space, the teacher evaluates the student’s response, identifies the strategic deficit, and produces a linguistic correction. This feedback serves as a ``gradient" that informs the iterative refinement of the guidance until behavioral convergence is achieved.

Formally, for a scenario with dialog context $x_s$, let
$y_T(x_s)$ denote the teacher’s response and
$y_S(x_s; g)$ denote the student’s response when conditioned on guidance $g$.
The objective is to identify a $g$ that minimizes a discrepancy measure:
\begin{equation}
    \min_{g} \; \ell\!\left(y_T(x_s),\, y_S(x_s; g)\right),
\end{equation}
where $\ell(\cdot)$ evaluates the difference between teacher and student behavior.

Rather than computing numerical gradients, GER elicits \emph{textual gradients}
through contrastive feedback:
\begin{equation}
\Delta g \;=\; \nabla_{\mathrm{text}}\!\left(y_T(x_s),\, y_S(x_s; g)\right),
\end{equation}
where $\Delta g$ is a natural-language feedback in the form of guidance describing how the student’s strategy
should be adjusted.

Guidance is refined iteratively by incorporating the feedback into the guidance:
\begin{equation}
g \leftarrow \mathcal{U}(g, \Delta g),
\end{equation}
until additional refinements yield no material improvement.
The resulting guidance captures the teacher’s strategic judgment in a persistent,
reusable textual form.

Importantly, this optimization occurs \emph{offline}.
Prompt optimization is used only to \emph{elicit} guidance during training.
At deployment, the student simply retrieves and applies the learned guidance, bypassing the
latency and computational cost associated with real-time reasoning models \citep{sui2025stop} or self-reflection loops (e.g., \citealt{renze2024self}).

\begin{figure}[h]  
  \centering
  \includegraphics[width=0.8\textwidth]{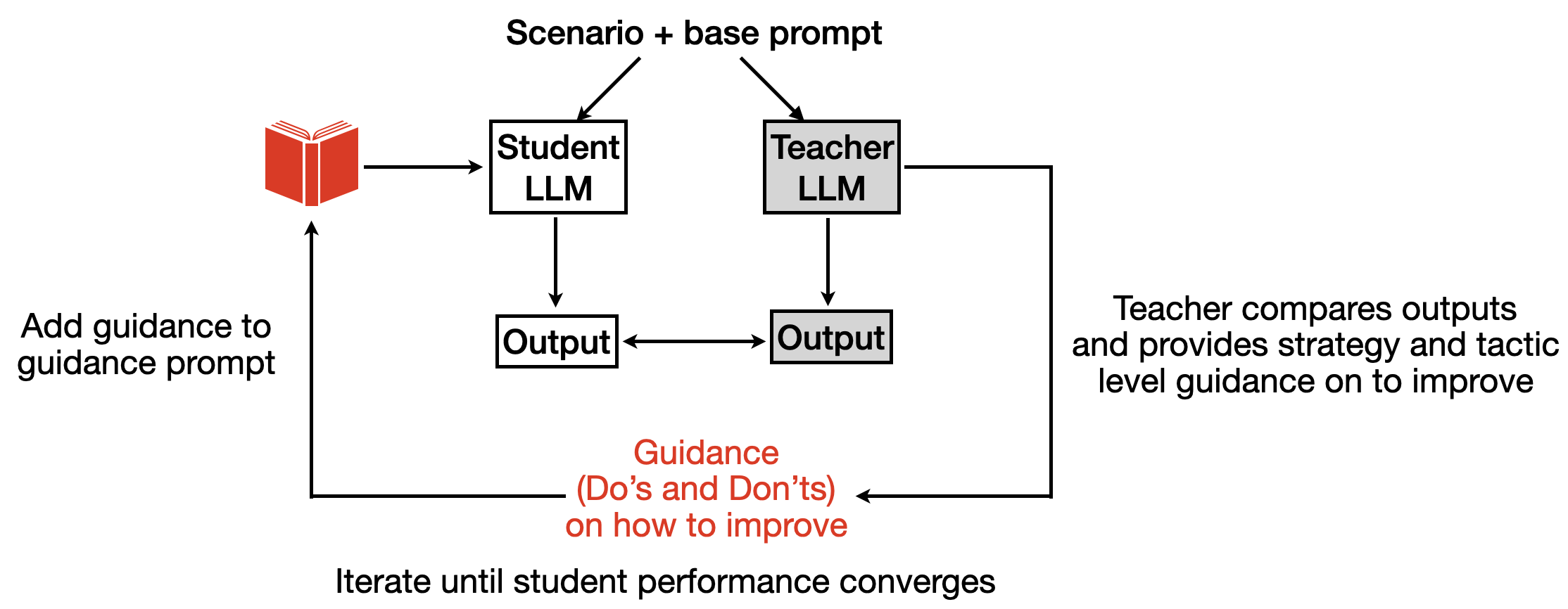}  
  \caption{Guidance learning via textual gradient.}
  \label{fig:strategy_teaching}
\end{figure}

Figure \ref{fig:strategy_teaching} illustrates the process of updating the guidance via textual gradient.

\subsection{Knowledge Transfer via Retrieval}\label{sec:retrieval}

Once optimized, the scenario-guidance pairs are stored explicitly in a library $\mathcal{L}$. 
At inference time, guidance for the the current dialog is obtained by finding the closest match scenarios (in terms of Euclidean distances) in the library and the associated guidance is appended to the student’s prompt, providing
state-contingent, closed-loop feedback that adapts behavior dynamically as the interaction
unfolds.

The retrieval-based knowledge transfer serves two conceptual purposes. First, it enables \emph{governance and modularity}.
Because guidance remains external to model parameters, it can be inspected, edited,
approved, or removed without retraining, supporting transparency and policy compliance. Second, it weakens the coupling between training data and deployment behavior.
Because guidance is applied selectively only when relevant, effective performance depends on
coverage of representative situations rather than exact replication of the full data
distribution.
This contrasts with fine-tuning, where spurious regularities may become permanently embedded
in parameters.

\paragraph{Scalability.}
GER operates through retrieval over a compact, domain-specific guidance library. In goal-oriented customer service settings, interactions are structured around a bounded set of intents, policies, and resolution pathways, so the number of strategically distinct interaction patterns is typically limited by the firm’s service design rather than by the space of possible utterances. The guidance library therefore stores reusable strategic policies rather than enumerating all possible dialogs, keeping its size manageable as deployment expands. 
Because guidance is retrieved from a fixed library rather than generated or retrained, the incremental cost per interaction depends primarily on usage volume rather than model complexity, making deployment costs predictable as scale increases. Together, these features allow deployment costs to scale predictably with usage, making the framework suitable for large-scale operational settings.


\subsection{Scenario Curriculum and Coverage} 
\label{sec:curriculum}

Because GER transfers knowledge through retrieval rather than parameter updates, its effectiveness
depends critically on whether the guidance library covers the conversational states the student is
likely to encounter at deployment. Guidance is elicited only for states observed during training;
states that are far from this support cannot benefit from distillation. Scenario construction is therefore
an intrinsic component of the learning algorithm rather than merely a data preprocessing step. In
this sense, the scenario library defines the \emph{support} over which retrieval-based distillation
operates.

\paragraph{Why coverage matters.}
Two properties distinguish scenario design in GER from conventional synthetic data generation and expert demonstrations.

First, multi-turn dialogs are sequential and path-dependent. If training trajectories are generated
solely by a teacher model, the resulting states reflect expert behavior and may omit precisely the
off-trajectory situations induced by imperfect student decisions at deployment.  Even small early deviations
by the student can alter the future course of the interaction, causing the learner to systematically
encounter states never observed during training. This endogenous distribution shift is well known in
imitation learning \citep{pomerleau1991efficient, ross2011reduction}. Consequently,
relying only on teacher- or human-generated conversations is insufficient.

Second, retrieval-based transfer changes the objective of data construction. In fine-tuning, synthetic
data must closely match the empirical distribution of human dialogs to avoid embedding spurious
regularities into parameters \citep{geirhos2020shortcut, bommasani2021opportunities, gui2023challenge, gao2025take}.
In GER, guidance is applied only when a nearby scenario exists. Effective deployment therefore
requires \emph{coverage} of the conversational state space rather than distributional matching. The
design goal is breadth of strategically distinct situations rather than aggregate realism.

\paragraph{Staged teacher--student transition.}
To achieve coverage of deployment-relevant states, we construct scenarios through a staged
teacher--student training curriculum. Intuitively, this curriculum introduces simpler, expert-led interactions
early in training and gradually increases the participation of the student in the generation as the process iterates and the student improves. Early on, the teacher
dominates dialog generation, producing coherent trajectories that provide stable and diagnostically
useful guidance. As the student acquires foundational behaviors through retrieving guidance from the current library $\mathcal{L}^{(b)}$, control progressively shifts toward
the student, exposing the system to more challenging and deployment-relevant states.

Formally, following interactive imitation learning \citep{ross2010efficient, ross2011reduction}, conversations are
generated in batches indexed by $b$. Each dialog is simulated for multiple turns and truncated at a
customer utterance to form a scenario,
\begin{equation}
\mathbf{x}^{(b,i)} =
(\mathbf{a}^{(b,i)}_1, \mathbf{c}^{(b,i)}_1, \ldots,
 \mathbf{a}^{(b,i)}_n, \mathbf{c}^{(b,i)}_n),
\end{equation}
where $\mathbf{a}^{(b,i)}_k$ and $\mathbf{c}^{(b,i)}_k$ denote the agent’s and customer’s utterances
at turn $k$.

Within each batch, agent turns are sampled from a mixture of the teacher and student models:
\begin{align}\label{eq:agent_mix}
\mathbf{a}^{(b,i)}_k =
\begin{cases}
y_{\mathcal{T}}(\mathbf{x}^{(b,i)}_{[:2k-1]}),
    & \text{with probability } p^{(b)}, \\
y_{\mathcal{S}}(\mathbf{x}^{(b,i)}_{[:2k-1]}, \mathcal{L}^{(b)}),
    & \text{with probability } 1 - p^{(b)},
\end{cases}
\end{align}
with a mixing schedule
\begin{equation}
p^{(b)} = \gamma^{b-1}, \text{ where } 0 < \gamma < 1.
\end{equation}

Thus $p^{(1)} = 1$, so early training relies entirely on teacher trajectories. As $b$ increases,
control progressively shifts toward the student, exposing the teacher to states induced by the
student’s evolving behavior. Newly elicited guidance is added to the library $\mathcal{L}^{(b)}$,
which in turn conditions subsequent student generations. This closed loop creates a curriculum in
which the learner first anchors on expert behavior and then gradually confronts harder,
deployment-relevant states—intuitively, ``learn to walk before you run.''

\paragraph{Coverage diagnostics and library expansion.}
Library construction proceeds iteratively rather than targeting a fixed dataset size. After each batch,
new scenarios are annotated with teacher-elicited guidance and added to the library, and student
performance is evaluated. We
continue expanding the library until marginal improvements plateau.

In practice, this stopping rule aligns closely with coverage diagnostics such as nearest-neighbor
distances between validation scenarios and library entries. When additional scenarios no longer
improve outcomes or reduce these distances, the library is deemed to have sufficient coverage. This
criterion controls growth while ensuring that retrieval remains both effective and computationally
lightweight.

\paragraph{Incorporating human dialogs.}
When real conversational logs are available, they are incorporated directly into the same framework to warm start the system.
Scenario dialogs can be extracted from observed interactions. These
examples are annotated with teacher-elicited guidance and added to the initial library before
synthetic expansion.

Because guidance is retrieved throughout both training and deployment, the influence of human
examples persists rather than being confined to initialization. In practice, a relatively small number
of high-quality human scenarios often suffices to anchor the library, with synthetic generation
expanding coverage around them. This hybrid strategy improves sample efficiency while maintaining
deployment robustness.

\section{Empirical Evaluation}\label{sec:experiment}

This section evaluates \textsc{GER} along the key dimensions articulated in
\S~\ref{sec:framework}.
We organize the empirical analysis around two overarching questions.
First, does guidance-based distillation via \textsc{GER} improve the performance and robustness of
student language models relative to standard alternatives?
Second, does the scenario curriculum in \S~\ref{sec:curriculum} support reliable deployment by
mitigating distribution shift in multi-turn dialog and ensuring adequate coverage of deployment-time
states?

We assess \textsc{GER} along five criteria:
(i) whether \textsc{GER} narrows the performance gap between student models and the teacher;
(ii) which dimensions of service quality improve;
(iii) whether guidance-based distillation outperforms parameter-based distillation via supervised
fine-tuning (SFT) and generic guidance;
(iv) whether \textsc{GER} remains effective for modern \emph{reasoning-capable} student LLMs, for which
test-time self-reflection can be effective but operationally costly; and
(v) whether distilled guidance generalizes across contexts and student models without retraining.

We then evaluate the scenario curriculum and coverage claims in \S~\ref{sec:curriculum}. Specifically,
we test (i) whether staged teacher--student transition during scenario construction mitigates
distribution shift at deployment; (ii) whether simulated scenarios provide adequate coverage of
human dialogs in embedding space; and (iii) whether augmenting simulated scenarios with limited human
dialogs yields incremental gains in live interactions.

We begin by describing the evaluation protocol.

\subsection{Evaluation Protocol}\label{sec:evaluation}

\paragraph{Teacher and student models.}
We use GPT-4\footnote{We used \texttt{gpt-4-0613}. For simplicity we will refer to it as GPT-4 in the remainder of the paper.} as the teacher throughout.
Our primary student models are LLaMA-2-7b-chat, LLaMA-2-13b-chat, LLaMA-2-70b-chat, and GPT-3.5-Turbo.
This set spans open-source and API-based models and covers a wide range of capacities.
To assess whether \textsc{GER} remains effective for modern reasoning-oriented systems, we additionally
evaluate \textsc{GER} on a set of reasoning-capable student LLMs in
\S~\ref{sec:reasoning_models_main} (model details and prompts appear in the Online Appendix).

\paragraph{Evaluation metrics.}
We assess dialog quality using three complementary approaches.

\emph{LLM-judged customer satisfaction.}
Following prior work \citep{li2024llms, gao2024llm}, each dialog is rated on a 1--5 satisfaction scale using few-shot prompting.
Unless otherwise noted, GPT-4 serves as the judge in the main text; robustness checks and prompts appear in the Online Appendix.
Judges are blinded to method condition and observe only the dialog transcript and the evaluation rubric.
To assess robustness to judge calibration, we evaluate results under two fixed prompt variants corresponding to relatively generous and relatively strict satisfaction standards. In both cases, the judge is provided with few-shot exemplars that anchor the scoring scale; conclusions are consistent across variants.

\emph{Static Human evaluation (Retrospective).}
To establish a human-centric baseline, we conduct systematic transcript reviews where independent human raters evaluate the recorded agent-customer interactions.
We conduct blind pairwise comparisons between student and teacher outputs to assess whether \textsc{GER} meaningfully closes the gap to the teacher.
Dialog order is randomized and left/right placement is counterbalanced to mitigate positional bias. 
We validate LLM-based satisfaction scores against human judgments and find strong agreement across multiple measures (full validation details in the Online Appendix). We also evaluate the multiple dimensions of agent conversations with human raters. 

\emph{Human-in-the-loop evaluation (Interactive).}
Participants role-play as customers and engage in live conversations with the agent via a custom web interface.
This first-person interactive setup captures subtleties---such as tone, responsiveness, and conversational flow---that retrospective evaluations may miss.

\paragraph{Presentation convention.}
To reduce redundancy across models and judge calibrations, we report both absolute satisfaction levels and
improvements relative to the base student (deltas) where helpful.
For LLM-judge tables, Panel~(a) uses the generous calibration and Panel~(b) uses the harsh calibration.

\subsection{Effectiveness and Robustness of \textsc{GER}}\label{sec:ger_effectiveness}

\subsubsection*{(i) Does \textsc{GER} Narrow the Gap to the Teacher?}

We first test whether \textsc{GER} narrows the gap between student models and the teacher in blind human
comparisons. For each teacher–student pair, we generate 32 independent customer–agent conversations under matched customer prompts. Each conversation is evaluated by a separate annotator, yielding 32 independent judgments per pair.

We recruited 256 participants in total. In each evaluation, participants are shown two anonymized dialogs—one generated by the teacher and one by the student—both interacting with the same customer LLM under identical initial conditions. Annotators are asked to select the better overall conversation. A 50\% selection rate corresponds to parity with the teacher.

\begin{table}[h!]
\caption{Human pairwise preference: percentage of times the student is selected over GPT-4 (teacher).}
\label{tab:human_pairwise}
\centering
\small
\renewcommand{\arraystretch}{0.9}
\begin{tabular}{lcccc}
\toprule
\textsc{Method} & \textsc{LLaMA-2-7b} & \textsc{LLaMA-2-13b} & \textsc{LLaMA-2-70b} & \textsc{GPT-3.5} \\
\midrule
\textsc{Base LLM} & 3\% & 25\% & 38\% & 22\% \\
\textsc{\Methodshort} & 34\% & 41\% & 41\% & 41\% \\
\midrule
\textsc{Gain (pp)} & +31 & +16 & +3 & +19 \\
\bottomrule
\end{tabular}
\end{table}

Table~\ref{tab:human_pairwise} shows that \textsc{GER} substantially improves all students relative to
their base performance, closing a meaningful fraction of the gap to GPT-4.
The largest gains occur for weaker students, but even higher-capacity models exhibit consistent
improvements.

Given the cost of large-scale human evaluation, we rely on LLM-judged satisfaction for the remaining
experiments and reserve human-in-the-loop evaluation for assessing scenario curation and deployment
robustness. Detailed examples illustrating the scenarios in which students benefit most from the guidance, as well as the specific guidance they receive, are provided in the online appendix.
\subsubsection*{(ii) Which Dimensions of Service Quality Improve?}\label{sec:rater_main}

Overall satisfaction does not identify which aspects of service quality improve.
We therefore evaluate four attributes drawn from the RATER framework \citep{parasuraman1988servqual}:
reliability, assurance, empathy, and responsiveness.
We omit tangibility because customer support interactions do not involve physical service cues.
Survey items and rating instructions appear in the Online Appendix.

Each agent configuration is evaluated by 13 to 23 participants.
Participants rate four multi-turn dialogs randomly selected from the test set, on each attribute, on a scale of 1 to 5.
Ratings are averaged at the dialog level and then aggregated by agent configuration.
To conserve space, Table~\ref{tab:rater_delta} reports $\Delta$ (GER minus Base); absolute levels and
additional details appear in the Online Appendix.

\begin{table}[h!]
\centering
\caption{Human attribute ratings: improvement from applying \textsc{GER} (GER $-$ Base).}
\label{tab:rater_delta}
\small
\renewcommand{\arraystretch}{0.9}
\begin{tabular}{lcccc}
\toprule
\textsc{Model} & $\Delta$ Reliability & $\Delta$ Assurance & $\Delta$ Empathy & $\Delta$ Responsiveness \\
\midrule
\textsc{LLaMA-2-7b}  & +0.58 & +0.62 & +0.82 & +0.66 \\
\textsc{LLaMA-2-13b} & +0.29 & $-$0.04 & +0.44 & +0.33 \\
\textsc{LLaMA-2-70b} & +0.23 & +0.19 & +0.12 & +0.11 \\
\textsc{GPT-3.5-Turbo}     & +0.71 & +0.96 & +0.27 & +0.60 \\
\bottomrule
\end{tabular}
\end{table}

\textsc{GER} improves all four dimensions for most models, with especially pronounced gains in empathy
and responsiveness.
Importantly, improvements occur across multiple attributes simultaneously, suggesting that guidance
corrects high-impact tactical errors rather than trading off one dimension against another.

\subsubsection*{(iii) Does \textsc{GER}'s Custom Guidance Outperform SFT and Generic Guidance?}
\label{sec:ger_vs_sft}

We next compare \textsc{GER} to parameter-based distillation via supervised fine-tuning (SFT).
To ensure comparability, \textsc{GER} and SFT use the same scenarios and teacher responses; the
difference is that \textsc{GER} elicits and retrieves guidance at inference time, whereas SFT embeds
behavior into model parameters.

We also compare \textsc{GER} to two generic guidance baselines.
\textsc{CoT Guidance} appends a chain-of-thought style rationale generated by the teacher without
observing the student’s output.
\textsc{Global Guidance} uses a single static ``handbook'' prompt applied uniformly across scenarios
and students. See the online appendix for the global guidance.

\begin{table}[h]
\caption{LLM-judged satisfaction (absolute) and improvement over the base student (deltas).}
\label{tab:main_compact}
\centering
\footnotesize
\renewcommand{\arraystretch}{0.9}

\begin{subtable}[t]{0.49\textwidth}
\centering
\caption{Generous Raters}
\begin{tabular}{lcccc}
\toprule
 & \multicolumn{3}{c}{\textsc{LLaMA-2}} & \textsc{GPT} \\
\cmidrule(lr){2-4} 
\textsc{Method} & \textsc{7B} & \textsc{13B} & \textsc{70B} & \textsc{3.5-Turbo} \\
\midrule
\textsc{Base LLM}        & 3.91 & 4.81 & 4.88 & 4.91 \\
\textsc{GER}             & {4.88} & 5.00 & 5.00 & 5.00 \\
\textsc{SFT}             & 4.63 & 4.94 & 5.00 & 4.68 \\
\textsc{CoT Guidance}    & 4.19 & 4.43 & 4.82 & 4.88 \\
\textsc{Global Guidance} & 4.44 & 4.50 & 4.75 & 5.00 \\
\midrule
$\Delta$\textsc{GER} (vs Base) & +0.97 & +0.19 & +0.12 & +0.09 \\
$\Delta$\textsc{SFT} (vs Base) & +0.72 & +0.13 & +0.12 & $-0.23$ \\
\bottomrule
\end{tabular}
\end{subtable}\hfill
\begin{subtable}[t]{0.49\textwidth}
\centering
\caption{Harsh Raters}
\begin{tabular}{lcccc}
\toprule
 & \multicolumn{3}{c}{\textsc{LLaMA-2}} & \textsc{GPT} \\
\cmidrule(lr){2-4} 

\textsc{Method} & \textsc{7b} & \textsc{13b} & \textsc{70b} & \textsc{3.5-Turbo} \\
\midrule
\textsc{Base LLM}        & 2.84 & 3.59 & 4.03 & 4.03 \\
\textsc{\Methodshort}    & 3.44 & {3.75} & {4.28} & {4.75} \\
\textsc{SFT}             & {3.75} & 3.47 & 4.06 & 3.81 \\
\textsc{CoT Guidance}    & 3.22 & 3.35 & 3.96 & 4.31 \\
\textsc{Global Guidance} & 3.31 & 3.44 & 3.88 & 3.75 \\
\midrule
$\Delta$\textsc{\Methodshort} (vs Base) & +0.60 & +0.16 & +0.25 & +0.72 \\
$\Delta$\textsc{SFT} (vs Base)          & +0.91 & $-$0.12 & +0.03 & $-$0.22 \\
\bottomrule
\end{tabular}
\end{subtable}
\end{table}

Table~\ref{tab:main_compact} reports absolute satisfaction and improvements over the base student.

\begin{figure}[h]
\centering
\includegraphics[width=\textwidth]{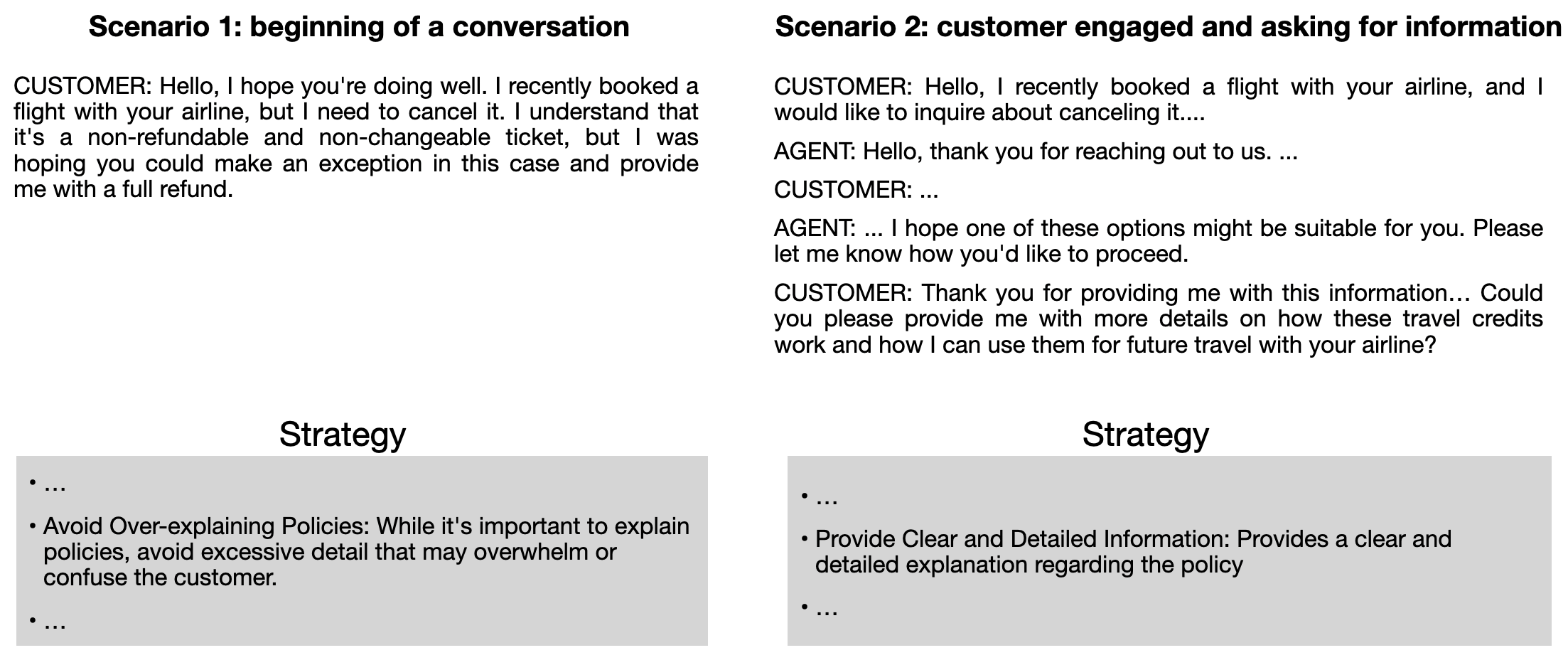}
\caption{Illustration: \textsc{GER} produces scenario-dependent guidance that can differ markedly across contexts.}
\label{fig:different_guidance}
\end{figure}

Overall, \textsc{GER} consistently improves student performance over the base model and is competitive
with or superior to SFT for higher-capacity students, while avoiding parameter updates.
Generic guidance underperforms \textsc{GER} under both harsh and generous raters, supporting the premise that
effective coaching requires guidance that is both \textit{targeted} (responding to observed student errors)
and \textit{contextual} (tailored to the scenario). Figure \ref{fig:different_guidance} illustrates how guidance can differ markedly across contexts for the same student, explaining why generic guidance is inadequate.

\subsubsection*{(iv) Does \textsc{GER} Help Reasoning-Capable Student LLMs?}
\label{sec:reasoning_models_main}

A natural question is whether guidance-based coaching remains useful for modern reasoning-capable
LLMs.
Such models can often improve their outputs via test-time self-reflection, but doing so increases
latency and inference cost---limitations that are especially salient in real-time customer service
settings.
\textsc{GER} is model-agnostic in the sense that both guidance elicitation and deployment rely solely on black-box querying of language models.
Scenario construction and guidance refinement are based on observed inputs and outputs and do not require access to model internals, reasoning traces, or architectural details.
As a result, \textsc{GER} is compatible with reasoning-capable student models while remaining independent of their internal deliberation mechanisms.

We apply \textsc{GER} to four reasoning LLMs with different capacities and sizes. See Online Appendix for more details. 

\begin{table}[h!]
\caption{Average customer satisfaction for reasoning-capable student LLMs.}
\label{tab:reasoning_main}
\centering
\footnotesize
\renewcommand{\arraystretch}{0.8}

\begin{subtable}[t]{0.49\textwidth}
\centering
\caption{Generous Raters}
\begin{tabular}{lcccc}
\toprule
\textsc{Method} & GLM-4.5 & DS-R1-7B & DS-R1-14B & DS-R1-32B \\
\midrule
\textsc{Base} & 4.84 & 4.75 & 4.00 & 4.44 \\
\textsc{GER}  & 5.00 & 4.88 & 4.63 & 4.66 \\
\midrule
$\Delta$ & +0.16 & +0.13 & +0.63 & +0.22 \\
\bottomrule
\end{tabular}
\end{subtable}\hfill\quad
\begin{subtable}[t]{0.49\textwidth}
\centering
\caption{Harsh Raters}
\begin{tabular}{lcccc}
\toprule
\textsc{Method} & GLM-4.5 & DS-R1-7B & DS-R1-14B & DS-R1-32B \\
\midrule
\textsc{Base} & 4.34 & 4.50 & 3.47 & 3.47 \\
\textsc{GER}  & 4.53 & 4.75 & 4.13 & 4.16 \\
\midrule
$\Delta$  & +0.19 & +0.25 & +0.66 & +0.69 \\
\bottomrule
\end{tabular}
\end{subtable}

\end{table}

Table~\ref{tab:reasoning_main} shows that \textsc{GER} improves satisfaction across all evaluated
reasoning-capable students.
Gains are largest for mid-strength distilled models, but remain positive even for the strongest
baseline, indicating that retrieved guidance complements rather than interferes with internal
reasoning.

Overall, \textsc{GER} offers a lightweight
alternative to slow test-time self-reflection: it improves response quality through reusable,
scenario-level guidance without increasing inference-time deliberation. Consistent with the operational motivation in the introduction, these results show that GER improves response quality even for reasoning-capable student models—without requiring additional inference-time deliberation.

\subsubsection*{(v) Does Distilled Guidance Generalize without Retraining?}\label{sec:transfer_main}
The preceding analyses show that guidance-based coaching improves performance across a wide range of student models, including reasoning-capable LLMs, in a focal context. We next examine a distinct and practically critical question: whether the distilled guidance possesses a meaningful degree of generalizability.

We evaluate generalization along two dimensions---across contexts and across student models---without retraining.

\paragraph{Cross-context transfer.}
We evaluate whether guidance distilled in one domain can effectively inform behavior in a distinct, yet structurally similar, service context. Specifically, we deploy a library developed for service ticket resolution to a ``lost luggage'' and request \$10{,}000 compensation.
While the surface content differs, the two domains share an underlying strategic structure: managing aggressive customer demands that exceed corporate policy.

\paragraph{Cross-model reuse.}
We reuse a library constructed for LLaMA-2-7b-chat and deploy it with other student models without rebuilding or retraining. This ``plug-and-play" capability is a unique advantage of \textsc{GER}; whereas Supervised Fine-Tuning (SFT) embeds knowledge into model-specific numerical weights, \textsc{GER} externalizes expertise into a model-agnostic textual form.




\begin{table}[h!]
\caption{Generalization without retraining: cross-context and cross-model transfer.}
\label{tab:transfer_context}
\centering
\footnotesize
\renewcommand{\arraystretch}{0.85}

\noindent
\makebox[0.49\textwidth]{\centering \textbf{Generous Raters}}
\hfill
\makebox[0.49\textwidth]{\centering \textbf{Harsh Raters}}
\vspace{-2em}

\begin{subtable}[t]{0.49\textwidth}
\centering
\caption{Cross-context (ticket $\to$ luggage)}
\begin{tabular}{lcccc}
\toprule
 & \multicolumn{3}{c}{\textsc{LLaMA-2}} & \textsc{GPT} \\
\cmidrule(lr){2-4} \cmidrule(lr){5-5}
\textsc{Method} & \textsc{7b} & \textsc{13b} & \textsc{70b} & \textsc{3.5-Turbo} \\
\midrule
\textsc{Base LLM}      & 4.50 & 5.00 & 5.00 & 5.00 \\
\textsc{\Methodshort} & 4.50 & 5.00 & 5.00 & 5.00 \\
\textsc{Fine-tuning}  & 5.00 & 5.00 & 4.88 & 5.00 \\
\bottomrule
\end{tabular}
\end{subtable}
\hfill
\begin{subtable}[t]{0.49\textwidth}
\centering
\caption{Cross-context (ticket $\to$ luggage)}
\begin{tabular}{lcccc}
\toprule
 & \multicolumn{3}{c}{\textsc{LLaMA-2}} & \textsc{GPT} \\
\cmidrule(lr){2-4} \cmidrule(lr){5-5}
\textsc{Method} & \textsc{7b} & \textsc{13b} & \textsc{70b} & \textsc{3.5-Turbo} \\
\midrule
\textsc{Base LLM}      & 3.13 & 4.25 & 4.50 & 4.63 \\
\textsc{\Methodshort} & 4.00 & 4.63 & 4.63 & 5.00 \\
\textsc{Fine-tuning}  & 4.50 & 4.63 & 4.63 & 4.75 \\
\bottomrule
\end{tabular}
\end{subtable}

\vspace{0.5ex}

\begin{subtable}[t]{0.49\textwidth}
\centering
\caption{Cross-model reuse (library built for 7b)}
\begin{tabular}{lcccc}
\toprule
 & \multicolumn{3}{c}{\textsc{LLaMA-2}} & \textsc{GPT} \\
\cmidrule(lr){2-4} \cmidrule(lr){5-5}
\textsc{Method} & \textsc{7b} & \textsc{13b} & \textsc{70b} & \textsc{3.5-Turbo} \\
\midrule
\textsc{Base LLM}        & 3.91 & 4.81 & 4.88 & 4.91 \\
\textsc{\Methodshort}   & 4.88 & 5.00 & 5.00 & 5.00 \\
\textsc{GER (Transfer)} & 4.88 & 5.00 & 4.88 & 5.00 \\
\bottomrule
\end{tabular}
\end{subtable}
\hfill
\begin{subtable}[t]{0.49\textwidth}
\centering
\caption{Cross-model reuse (library built for 7b)}
\begin{tabular}{lcccc}
\toprule
 & \multicolumn{3}{c}{\textsc{LLaMA-2}} & \textsc{GPT} \\
\cmidrule(lr){2-4} \cmidrule(lr){5-5}
\textsc{Method} & \textsc{7b} & \textsc{13b} & \textsc{70b} & \textsc{3.5-Turbo} \\
\midrule
\textsc{Base LLM}        & 2.84 & 3.59 & 4.03 & 4.03 \\
\textsc{\Methodshort}   & 3.44 & 3.75 & 4.28 & 4.75 \\
\textsc{GER (Transfer)} & 3.44 & 3.91 & 3.94 & 4.44 \\
\bottomrule
\end{tabular}
\end{subtable}

\end{table}

The results in Table~\ref{tab:transfer_context} establish the generalizability of \textsc{GER} as a robust strategic asset. The cross-context success suggests that distilled guidance captures ``portable" service principles—such as de-escalation and staged remedies—that transcend specific domains to manage the underlying structure of high-stakes interactions. Furthermore, the cross-model findings substantiate that this expertise is model-agnostic. By externalizing judgment into a textual form, \textsc{GER} avoids the ``parameter lock" of traditional fine-tuning, allowing firms to future-proof their operations by instantly ``hot-swapping" the same strategic library across evolving model architectures.

\subsection{Effectiveness of Scenario Curation}\label{sec:scenario_curation_eval}

This section evaluates whether the scenario curriculum in \S~\ref{sec:curriculum} produces guidance
libraries that (i) remain well aligned with deployment-time student behavior, mitigating distribution
shift in multi-turn dialogs, (ii) provide adequate coverage of human interactions when libraries are
constructed primarily from simulated dialogs, and (iii) offer incremental value when augmented with human-generated dialogs.
These tests correspond directly to the two design challenges emphasized in \S~\ref{sec:curriculum}:
sequential distribution shift and coverage adequacy under retrieval-based transfer.

\subsubsection*{(i) Does Staged Teacher-Student Transition in Scenario Curation Mitigate Distribution Shift?}
\label{sec:distribution_shift_main}

We compare our scenario curation with a staged teacher-student transition to an ablated
\emph{teacher-only} variant, which serves as an ablation to mimic the logic of classic behavioral cloning. In this baseline, all scenarios are generated solely from
teacher--customer interactions and the student never contributes to dialog generation. This approach represents learning from static expert demonstrations without accounting for the learner's own behavioral distribution.
Both libraries contain the same number of scenarios and use the same retrieval mechanism and
inference-time prompting pipeline; the only difference is whether the student participates during
scenario construction.

\paragraph{Outcome-based evidence.}
Table~\ref{tab:ds_eval_compact} reports GPT-4--judged customer satisfaction for dialogs generated
using guidance retrieved from each library.
Across all student models and under both judge calibrations, the teacher-only library consistently
underperforms the mixed-agent library.
In several cases, performance approaches---or falls close to---that of the base student model,
indicating that guidance derived solely from expert trajectories is less effective when applied to
student-generated interactions at deployment.

\begin{table}[h!]
\caption{Mixed-agent curation vs.\ teacher-only curation: LLM-judged satisfaction.}
\label{tab:ds_eval_compact}
\centering
\footnotesize
\renewcommand{\arraystretch}{0.9}

\begin{subtable}[t]{0.49\textwidth}
\centering
\caption{Generous Raters}
\begin{tabular}{lcccc}
\toprule
 & \multicolumn{3}{c}{\textsc{LLaMA-2}} & \textsc{GPT} \\
\cmidrule(lr){2-4} \cmidrule(lr){5-5}
\textsc{Method} & \textsc{7b} & \textsc{13b} & \textsc{70b} & \textsc{3.5-Turbo} \\
\midrule
\textsc{Base LLM} & 3.91 & 4.81 & 4.88 & 4.91 \\
\textsc{\Methodshort} & {4.88} & 5.00 & 5.00 & 5.00 \\
\textsc{\Methodshort (teacher-only)} & 4.50 & 4.75 & 4.63 & 4.81 \\
\bottomrule
\end{tabular}
\end{subtable}\hfill
\begin{subtable}[t]{0.49\textwidth}
\centering
\caption{Harsh Raters}
\begin{tabular}{lcccc}
\toprule
 & \multicolumn{3}{c}{\textsc{LLaMA-2}} & \textsc{GPT} \\
\cmidrule(lr){2-4} \cmidrule(lr){5-5}
\textsc{Method} & \textsc{7b} & \textsc{13b} & \textsc{70b} & \textsc{3.5-Turbo} \\
\midrule
\textsc{Base LLM} & 2.84 & 3.59 & 4.03 & 4.03 \\
\textsc{\Methodshort} & {3.44} & {3.75} & {4.28} & {4.75} \\
\textsc{\Methodshort (teacher-only)} & 3.25 & 3.56 & 4.06 & 4.38 \\
\bottomrule
\end{tabular}
\end{subtable}
\end{table}

\paragraph{Retrieval-level evidence (distribution shift proxy).}
To quantify how well each library represents deployment-time interactions, we compute the 
Euclidean distance from the embedding of each test scenario to its nearest neighbor in the library.
Smaller distances indicate that retrieval can identify more closely aligned precedents and,
consequently, supply more relevant guidance.
Table~\ref{tab:distance_compact} shows that the teacher-only library increases nearest-neighbor
distances for every student model, with increases of up to 11.6\%.
These results indicate systematic misalignment between teacher-only libraries and the conversational
states realized by student models at deployment.

\begin{table}[h!]
\caption{Distribution shift proxy: average distance from a test scenario to its closest library scenario.}
\label{tab:distance_compact}
\centering
\small
\renewcommand{\arraystretch}{0.9}
\begin{tabular}{lcccc}
\toprule
 & \multicolumn{3}{c}{\textsc{LLaMA-2}} & \textsc{GPT} \\
\cmidrule(lr){2-4} \cmidrule(lr){5-5}
\textsc{Library} & \textsc{7b} & \textsc{13b} & \textsc{70b} & \textsc{3.5-Turbo} \\
\midrule
\textsc{\Methodshort} & 0.179 & 0.172 & 0.181 & 0.169 \\
\textsc{\Methodshort (teacher-only)} & 0.192 & 0.192 & 0.191 & 0.174 \\
\midrule
\textsc{\% Increase} & 7.3\% & 11.6\% & 5.5\% & 3.0\% \\
\bottomrule
\end{tabular}
\end{table}

\subsubsection*{(ii) Do Simulated Scenarios Provide Adequate Coverage of Human Dialogs?}
\label{sec:coverage_human_main}

Because the guidance library is queried at inference time, its usefulness depends on whether
simulated scenarios span the same semantic regions as real human interactions. We therefore compare
simulated and human dialogs in the same embedding space used for retrieval.

\paragraph{Human data collection.}
We recruited 136 participants via Prolific to role-play passengers attempting to cancel restricted,
non-refundable tickets. Participants interacted with either LLaMA-2-7b-chat or GPT-4 through a
lightweight chat interface, yielding 136 dialogs comprising 1{,}302 utterances.

\paragraph{Coverage results.}
Across multiple coverage diagnostics (detailed in Appendix~\ref{app:epsilon_ball}), simulated
libraries cover the large majority of human scenarios. Uncovered cases are concentrated primarily in
idiosyncratic openings that reflect the inherent unpredictability of real users. Consistent
with the goal-oriented nature of the task, dialogs quickly converge toward common interaction
patterns, and coverage increases rapidly after the first few turns.

Qualitative visualizations (Appendix~\ref{app:coverage}) show that simulated and human scenarios
occupy the same semantic regions of the embedding space, with simulation largely expanding coverage
by interpolating within plausible behaviors rather than introducing qualitatively new modes.

These findings indicate that simulated scenarios capture the dominant behavioral
structure relevant for retrieval and guidance. In practice, simulation provides broad baseline
coverage, while human-authored dialogs mainly contribute incremental improvements by filling a small
number of edge cases.

\subsubsection*{(iii) Does Augmenting Simulated Scenarios with Human Dialogs Add Value?}
\label{sec:human_aug_main}

We next examine whether libraries constructed primarily from simulated scenarios degrade perceived
quality in live interactions and whether augmenting them with human dialogs yields incremental gains.

\paragraph{Human-in-the-loop evaluation.}
In this experiment, we use the student LLM is GPT-3.5-Turbo. We compare its performance after distillation using \textsc{GER} with and without human dialogs. To evaluate performance, we recruit 132 participants to engage in live conversations with GPT-3.5-Turbo acting as the agent. Each participant completes at least four conversational turns and then reports their satisfaction on a 1--5 scale. We evaluate GPT-3.5-Turbo under two library settings: (i) a simulated-only library containing 684 scenarios, and (ii) an augmented library consisting of the same 684 simulated scenarios plus an additional 606 human-derived scenarios.

The augmented library yields a higher mean satisfaction score (4.32 vs.\ 4.13), a 0.19-point increase on the five-point scale.
While this difference is not statistically distinguishable from zero in our sample ($p=0.44$), the distribution shifts in the predicted direction: the augmented condition exhibits fewer low ratings and a modest rightward shift toward higher satisfaction.
These patterns suggest that simulated scenarios already cover common conversational states, while human dialogs provide incremental gains by reducing occasional failure cases.

\paragraph{Implication.}
Taken together, the results indicate that guidance-based distillation can rely primarily on simulated scenario generation for broad behavioral coverage, with smaller amounts of human data serving to seed or validate edge cases. This hybrid strategy substantially improves data efficiency while maintaining comparable deployment-time performance.

\subsection{Discussion}\label{sec:discussion}

Our findings suggest that guidance-based distillation represents a distinct paradigm for improving weaker language models. Rather than embedding behavior directly into parameters through fine-tuning, \textsc{GER} externalizes strategic knowledge into a reusable guidance library that is retrieved at inference time. By decoupling knowledge from model weights, training becomes a problem of covering decision-relevant states, while deployment becomes a problem of selectively retrieving and updating guidance rather than re-optimizing parameters. 

This has two implications. First, effective transfer depends on \emph{coverage} rather than distributional matching: the library must span the conversational states the student is likely to encounter, not replicate the empirical distribution of dialogs. Staged teacher--student scenario construction improves performance by aligning guidance with student-induced trajectories. Second, retrieval enables selective adaptation. Because guidance is applied only when contextually relevant, corrections can be targeted, and the rest of the library remains reusable across models and contexts.

These implications have practical advantages. Storing strategies externally allows behavior to be inspected, revised, or governed without retraining, supporting rapid policy updates and low-latency deployment. As a result, \textsc{GER} is particularly attractive in high-volume or resource-constrained settings where stronger models are costly to serve but smaller models benefit from targeted coaching.

At the same time, guidance-based distillation is not a universal substitute for parameter-level adaptation. Its effectiveness depends on both adequate library coverage and a student with sufficient baseline capability to interpret and execute retrieved guidance. Because \textsc{GER} transfers high-level strategic intent rather than low-level linguistic patterns, it is less suited to tasks that require precise stylistic or structural replication, where dense statistical regularities are more naturally embedded through fine-tuning. We therefore view \textsc{GER} as complementary to existing approaches: a modular mechanism for managing strategic behavior through guidance, alongside parameter-based methods for domain-specific linguistic alignment.

\section{Conclusion}\label{sec:conclusion}

We introduce \textsc{GER}, a guidance-based distillation framework that improves smaller language
models by transferring strategic guidance from stronger models through inference-time coaching
rather than parameter updates. By eliciting, storing, and retrieving scenario-level guidance,
\textsc{GER} enables deployable models to inherit much of the competence of larger systems while
preserving efficiency, portability, and operational control.

Empirically, guidance-based coaching consistently improves performance across student
architectures and evaluation settings, often matching the gains of fine-tuning without requiring
retraining. These results highlight a broader principle: in interactive, sequential decision
problems, performance gains need not rely exclusively on modifying model weights. Strategic
knowledge can instead be externalized and selectively applied at inference time, enabling targeted,
reusable adaptation across contexts and models.

Conceptually, our work situates guidance-based distillation within two complementary paradigms for
improving language models in applied settings: parameter adaptation through fine-tuning and
behavioral shaping through prompt-based methods. While fine-tuning remains the dominant approach
in marketing applications and a natural benchmark for distillation, emerging research shows that
prompts can function as optimized carriers of reasoning, strategy, and behavioral control. Our
contribution is to operationalize this philosophy in a marketing context by developing a
systematic, retrieval-based distillation procedure tailored to multi-turn, strategic interaction.

Customer service provides a canonical setting for this approach. Interactions are path-dependent
and strategically complex, making static parameter updates vulnerable to distribution shift and
insufficient for governing nuanced behaviors such as de-escalation, remedy sequencing, and policy
enforcement. In such environments, organizations must balance performance with cost, latency, and
control. Guidance-based distillation offers a computationally light and auditable mechanism for navigating
these trade-offs, while remaining compatible with existing deployment pipelines.

Although we focus on customer service, the underlying logic extends to other marketing and
decision environments characterized by repeated interaction, path dependence, and governance
constraints. More broadly, \textsc{GER} is well suited to settings where performance gaps exist
between a weaker and a stronger model on a focal task, and organizations must balance quality
with cost, latency, and control. The framework relies on the student model possessing sufficient
instruction-following and reasoning capabilities to internalize and execute retrieved guidance.
As general-purpose LLMs continue to improve—particularly in their ability to follow structured
instructions—this prerequisite becomes increasingly easy to satisfy. At the same time,
task-specific performance differences across models are likely to persist, ensuring the ongoing
relevance of adaptation mechanisms such as \textsc{GER}. 

Future research may explore adaptive curriculum design for scenario curation, such as a more adaptive schedule tied to the student’s improvement
trajectory or to discounted future performance gains. We also see ample opportunity to extend our framework to other applications of human-AI interactions. More broadly, we hope this work encourages continued study of
prompt-based learning as practical, controllable complements to
parameter-based adaptation.

\section*{Funding and Competing Interests}
All authors certify that they have no affiliations with or involvement in any organization or entity with any financial interest or non-financial interest in the subject matter or materials discussed in this manuscript.
This project was supported by a National Science Foundation Grant.





\FloatBarrier
\newpage
{
\bibliographystyle{informs2014} 

\bibliography{Neurips2024/LLMref}}

\newpage 
\begin{APPENDIX}{}


\section{GER Implementation and Validation Details}

This appendix provides concise implementation and validation details that support the interpretation and credibility of our results. Additional robustness checks, expanded ablations, qualitative examples, and diagnostic analyses are deferred to the Online Appendix.

\subsection{Validation of LLM-Based Outcome Measures}

Several of our outcome measures, including customer satisfaction and issue resolution quality, are assessed using LLM-based evaluators. To validate these measures, we conducted a human–LLM agreement study in which independent human annotators evaluated a subset of dialogs using the same criteria as the LLM judge.

Across all evaluated dimensions, human–LLM agreement is high. Correlations between average human scores and LLM scores is 0.900 (Pearson) and 0.806 (Spearman), with intraclass correlation coefficients of 0.772. These results indicate that LLM-based evaluations closely track human judgments in this domain. Full details of the evaluation protocol, prompts, annotator instructions, and robustness checks are provided in the Online Appendix \S 2.

\subsection{Global Guidance Baseline}

In addition to scenario-specific guidance, we construct a \emph{Global Guidance} baseline consisting of high-level best-practice principles distilled from expert dialogs. Global guidance is intended to capture general norms of effective customer service without conditioning on specific dialog states.

Examples of global guidance include:
\begin{itemize}
    \item Maintain a polite and empathetic tone throughout the interaction.
    \item Clearly acknowledge the customer’s concern before proposing solutions.
    \item Avoid making commitments that violate stated policy constraints.
\end{itemize}

While global guidance improves performance relative to no guidance, it lacks the situational specificity needed to address the strategic challenges of multi-turn interactions. The complete global guidance handbook and construction procedure are provided in the Online Appendix \S 7.

\section{Scenario Coverage and Distribution Diagnostics}

This section provides additional diagnostics assessing whether the guidance library adequately covers the space of dialog scenarios encountered at inference time.

\subsection{$\epsilon$-Ball Coverage}\label{app:epsilon_ball}

\paragraph{Interpretation of coverage metrics.}
Because \textsc{GER} retrieves guidance by nearest-neighbor matching in the scenario embedding space,
effective deployment requires that evaluation scenarios lie sufficiently close to at least one library
scenario. Coverage of the embedding space---rather than distributional matching of dialog statistics---is therefore the relevant notion of adequacy. We compute $\varepsilon$-ball coverage metrics as a diagnostic to assess whether there are gross mismatches between dialog states represented in the guidance library and those encountered at deployment.

\subsubsection*{Do simulated scenarios cover human scenarios?}

To quantify the extent to which simulated scenarios cover the space of human dialogs, we introduce an \emph{asymmetric $\varepsilon$-ball recall} metric inspired by set-distance and coverage analyses \citep{dubuisson1994modified, huang2005coverage}.
Let $\mathcal{D}_{\text{human}}$ and $\mathcal{D}_{\text{sim}}$ denote the sets of human and simulated scenario embeddings.
A human scenario $x \in \mathcal{D}_{\text{human}}$ is said to be \emph{covered} if there exists a simulated scenario $z \in \mathcal{D}_{\text{sim}}$ such that $\lVert x - z \rVert_2 \le \varepsilon$.
We report the directional recall
\[
\%\texttt{Covered}(\text{human} \to \text{sim})
= \frac{1}{|\mathcal{D}_{\text{human}}|}
\sum_{x \in \mathcal{D}_{\text{human}}}
\mathbf{1}\{\exists z \in \mathcal{D}_{\text{sim}} : \lVert x - z \rVert_2 \le \varepsilon\},
\]
and analogously $\%\texttt{Covered}(\text{sim} \to \text{human})$.
This measure is asymmetric and directional: it evaluates whether one set provides adequate support for another rather than testing bidirectional similarity.

Because coverage increases monotonically with $\varepsilon$, we evaluate recall as a function of $\varepsilon$ and report results at the smallest $\epsilon$ for which $\%\texttt{Covered}(\text{sim} \to \text{human}) = 100\%$.
At this threshold, every simulated scenario lies within the semantic support of observed human interactions; meanwhile, $\%\texttt{Covered}(\text{human} \to \text{sim})$ reaches approximately 85\% (Figure~\ref{fig:coverage_subplots}a), implying that most human scenarios have nearby simulated counterparts.
\begin{figure}[h!]
  \centering
  \begin{subfigure}[b]{0.35\textwidth}
    \centering
    \includegraphics[width=\textwidth]{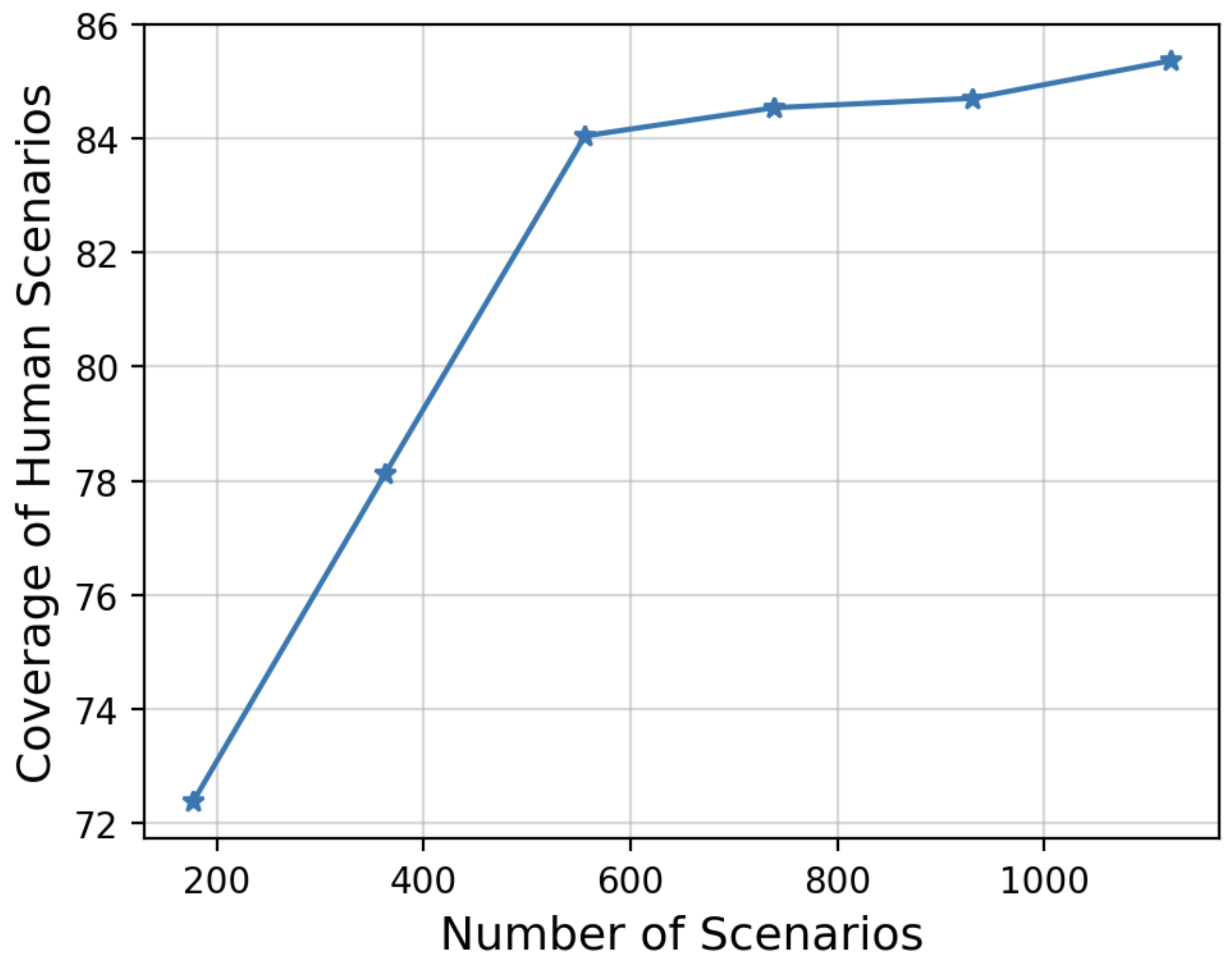}
  \end{subfigure}
  \hspace{1.5cm}
  \begin{subfigure}[b]{0.35\textwidth}
    \centering
    \includegraphics[width=\textwidth]{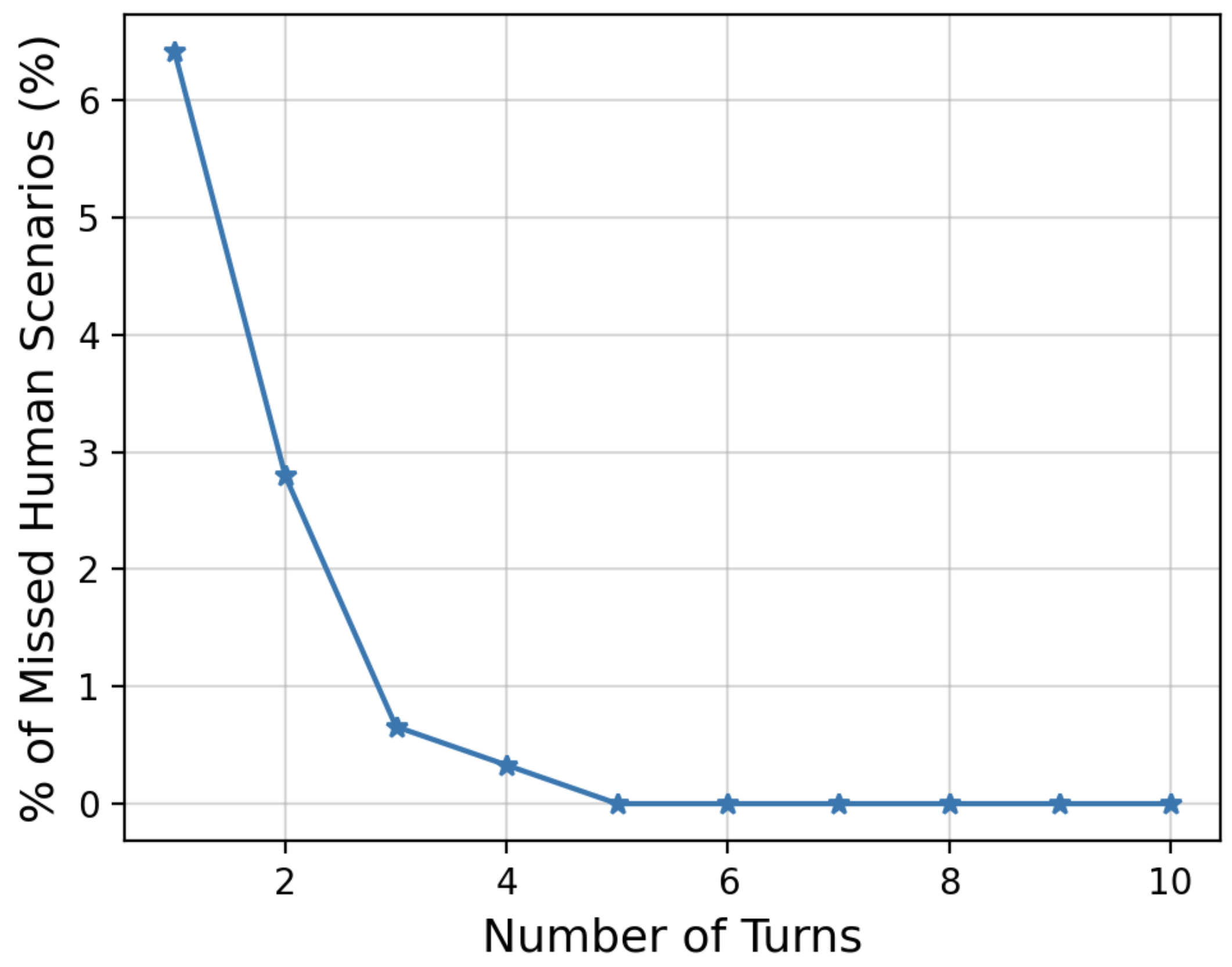}
  \end{subfigure}
  \caption{Coverage metrics as the number of synthetic scenarios increases. (a) The percentage of human-generated scenarios covered by simulated scenarios vs. the number of simulated
  scenarios. (b) Distribution of the number of turns in human scenarios that are not represented by LLM-simulated scenarios.}
  \label{fig:coverage_subplots}
\end{figure}
The remaining uncovered cases (approximately 15\%) are concentrated in short, idiosyncratic openings that reflect the variability of real users.
Consistent with the goal-oriented nature of the task, dialogs rapidly converge toward common interaction patterns after the initial turns; consequently, later-stage states exhibit substantially higher coverage.
Thus, simulated scenarios capture the dominant semantic regions relevant for retrieval and guidance, with residual gaps largely confined to early conversational phases.

Overall, these diagnostics suggest that simulation expands coverage primarily by interpolating within the space of plausible human behaviors rather than extrapolating beyond it, supporting the use of simulated scenarios as an efficient mechanism for constructing retrieval libraries.

\subsection{Coverage Visualization Diagnostics}\label{app:coverage}

To provide an intuitive view of how well the guidance library spans the space of dialog states
encountered at deployment, we visualize scenario embeddings using a two-dimensional UMAP
projection.\footnote{UMAP provides a nonlinear low-dimensional embedding that preserves local
neighborhood structure but may distort global distances. We therefore treat these visualizations as
descriptive rather than quantitative evidence of coverage.}

\begin{figure}[ht]
    \centering
    \includegraphics[width=0.8\textwidth]{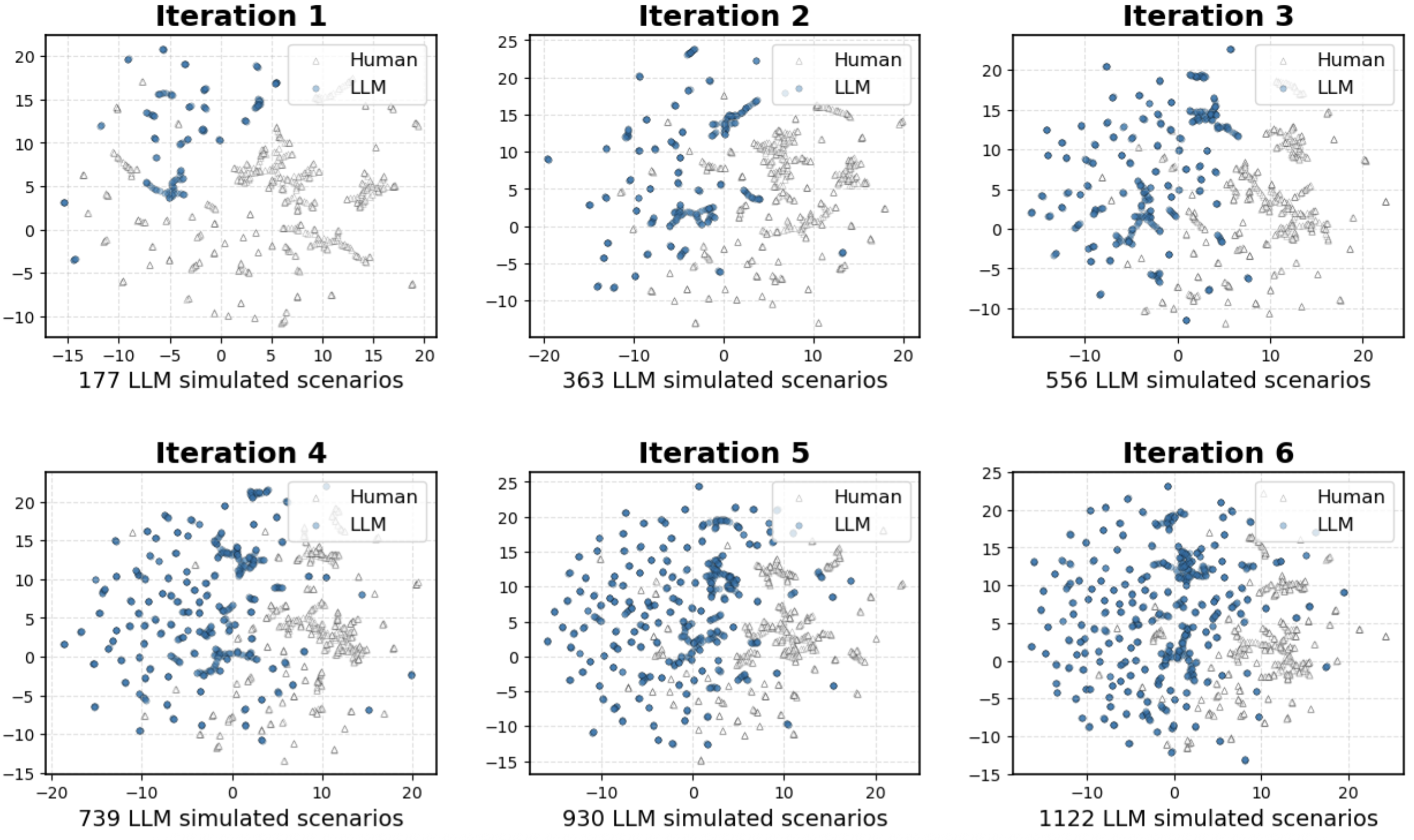}
    \caption{UMAP visualization of dialog scenarios. Training (library) and evaluation scenarios
    exhibit substantial overlap, with no large isolated regions unique to evaluation data,
    suggesting broad support of the deployment state space by the guidance library.}
    \label{fig:umap}
\end{figure}

Figure~\ref{fig:umap} provides a qualitative illustration of the embedding geometry underlying
retrieval in \textsc{GER}. Evaluation scenarios are largely interspersed with training scenarios,
rather than forming separate clusters or out-of-distribution regions. This pattern suggests that
deployment interactions lie within the same semantic support as library scenarios, consistent with
the premise that nearest-neighbor retrieval can supply relevant guidance rather than relying on
memorization of a small number of isolated cases.

\newpage
    \renewcommand{\thesection}{OA \arabic{section}}
    \renewcommand{\thesubsection}{\thesection.\arabic{subsection}}
    \setcounter{section}{0}
\section*{Online Appendix}

\noindent\textit{This online appendix provides additional implementation details, diagnostics, and qualitative analyses intended to support replication and to help readers understand how the proposed method behaves in practice. The material here is included to clarify design choices, assess robustness, and illustrate mechanisms. All core findings and conclusions of the paper can be understood without reference to this appendix.}


\section{Prompts used in training}
\textbf{$\Delta_\text{text}$} instantiated with an LLM that compares the student's response with the teacher's response and provides feedback. The prompt used in $\Delta_\text{text}$ is

\begin{tcolorbox}[
  colback=gray!15,   
  boxrule=0pt,       
  frame hidden,      
  breakable,
  left=6pt,right=6pt,top=6pt,bottom=6pt
]
\begin{Verbatim}[
  fontsize=\footnotesize,
]
You are a customer-support manager at an airline.  Compare the replies of two agents 
to the *same* customer request:
• **+AGENT (Good)** : the exemplary standard  
• **+AGENT (Bad)**  : needs coaching
────────────────────────────────────
CONVERSATION FORMAT
────────────────────────────────────
SCENARIO:
AGENT: Hello, this is agent 06520. How may I help you?  
CUSTOMER: <customer’s query>  
AGENT: <agent’s response>  
... (back-and-forth continues) ... 
CUSTOMER: <final query identical for both agents>

+AGENT (Good):  
AGENT: <good agent’s reply>

+AGENT (Bad):  
AGENT: <bad agent’s reply>
────────────────────────────────────
TASK  
1. **Compare the Good and Bad replies** on:  
    • communication strategy and tactics 
    • content accuracy and coverage
    • tone & empathy
    • structure / flow 
2. **Draft a “Guideline List”** for the Bad agent that will help her match the Good
   agent in this specific scenario.
   For each guideline:  
   • State the what the agent should do or should avoid.  
   • Provide an example from the Good agent's for the Bad agent to emulate.
   • In addition to your observation, there is also some feedback from other 
   customers why they believe the Bad agent's response is inadequate. Incorporate the 
   feedback in the guideline.
   ────────────────────────────────────
    FEEDBACK
   ────────────────────────────────────
    {explanation}
    
3. If the two replies are already equivalent on all key aspects, leave the Guideline 
List empty.

OUTPUT FORMAT  
Produce **only** the “Guideline List:” section—nothing else.
\end{Verbatim}
\end{tcolorbox}

The prompt used in the LLM to act as $\ell(\cdot, \cdot)$ is:
\begin{tcolorbox}[
  colback=gray!15,   
  boxrule=0pt,       
  frame hidden,      
  breakable,
  left=6pt,right=6pt,top=6pt,bottom=6pt
]
\begin{Verbatim}[fontsize = \footnotesize]
You are an impartial evaluator acting as the customer in the scenario below.
Rate the agent’s based on their interaction with the customer below. Rate **solely by 
how closely it matches the “excellent” reference reply**—in content, tone, structure, 
and customer-service strategy.
────────────────────────────────────
AGENT - CUSTOMER INTERACTION
────────────────────────────────────
"{student_reply}"
────────────────────────────────────
REFERENCE INTERACTION (benchmark = “excellent”)
────────────────────────────────────
{teacher_reply}

RATING SCALE (choose ONE word)  
unacceptable | acceptable | good | very good | excellent

GUIDELINES  
• **Content accuracy & coverage** – Does it convey the same solutions / information?  
• **Tone & empathy** – Is politeness and warmth comparable?  
• **Structure & flow ** – Does it follow a similar flow?  
• **Customer-service strategy** – Does it use the same problem-solving approach?  

OUTPUT FORMAT  
Return one rating word from the scale above in lower-case and an explanation for the
rating, in the following format:
[Rating]: one word rating here
[Explanation]: explanation here 

For example:
[Rating]: good
[Explanation]: The agent failed to ...
\end{Verbatim}
\end{tcolorbox}

\section{Implementation Details and Diagnostics}

This section provides additional information on the LLM-based evaluation procedures used to assess customer satisfaction, issue resolution quality, and related outcome measures.
\subsection{Evaluation Protocol}

LLM judges are prompted to provide an overall rating based on the agent-customer interactions. 
\begin{tcolorbox}[
  colback=gray!15,   
  boxrule=0pt,       
  frame hidden,      
  breakable,
  left=6pt,right=6pt,top=6pt,bottom=6pt
]
\begin{Verbatim}[
  fontsize=\footnotesize,
]
Role-play as a customer evaluating satisfaction of a customer support agents
based on the interactions. Please rate the agent in terms of the quality of 
his/her interaction. Give your rating as one of the five levels: 1: 
unacceptable, 2: acceptable, 3: good, 4: very good, 5: excellent.

As a benchmark of quality of interaction, the agent is rated as excellent in 
the following conversations. It is important to note that the agent might not
be able to provide a solution that completely satisfies the customer due to
the constraint of company policy.  			

{example 1}

{example 2}

Show your answer in the following format: 			
+Rating: [rating here].			
+Explanation: [explanation here]
\end{Verbatim}
\end{tcolorbox}

\subsection{Human--LLM Agreement and Robustness}\label{app:agreement}

To assess the reliability of LLM-based evaluations, we conducted a human–LLM agreement study using independent human annotators. Annotators were provided with the same dialog transcripts and evaluation criteria as the LLM judge.

We study two aspects of agreement:
\begin{enumerate}
    \item Inter-rater reliability between two human raters.
    \item Agreement between the average of all human ratings and the LLM rating.
\end{enumerate}

We consider four common evaluation metrics: Cohen’s kappa \citep{cohen1960coefficient}, Pearson correlation, Spearman correlation, and the intraclass correlation coefficient (ICC) \citep{koo2016guideline}. These metrics capture different aspects of agreement:  
\textit{Cohen’s kappa} corrects for chance agreement in categorical ratings, while \textit{Pearson} and \textit{Spearman correlations} measure linear and rank-based consistency, respectively. The \textit{ICC} captures absolute agreement in ratings and is widely used for assessing rater consistency on numerical scales.

\begin{table}[ht]
\centering
\caption{Rating agreement}
\label{tab:new}\small
\begin{tabular}{lcccc}
\toprule
 & Pearson & Spearman & ICC & Cohen's $\kappa$ \\
\midrule
 Human-Human (avg)
& 0.807 
& 0.769 
& 0.732 
& 0.302 \\
Avg Human - LLM (GPT-4) & 0.900 & 0.806 & 0.772 & --\\
\bottomrule
\end{tabular}
\end{table}

To obtain these metrics, we conducted a human evaluation study involving six human raters, each of whom rated the same ten conversations. We removed two respondents who spent significantly shorter time (less than half of the other respondents) and keep three respondents who spent 10.5 min on average.  GPT-4 was also used to rate these conversations, resulting in four total ratings per conversation.

\paragraph{Human--Human Agreement.}
The agreement across human raters is high across all four metrics— Pearson correlation (0.807), Spearman correlation (0.769), ICC (0.732) and Cohen’s kappa (0.302)—indicating strong consistency in raters’ evaluations. In particular, the high Pearson, Spearman, and ICC values reflect substantial agreement in relative and absolute judgments across conversations, while the moderate Cohen’s kappa suggests that most remaining disagreement arises from differences in exact category assignments on the discrete rating scale rather than from divergent underlying assessments.

\paragraph{LLM alignment with human consensus.}
We compare the LLM’s ratings to the average human rating for each conversation and find a high degree of alignment, with a Pearson correlation of 0.900, a Spearman correlation of 0.806, and an ICC of 0.772. The high Pearson correlation indicates that the LLM closely matches the variation in human consensus across conversations, while the Spearman correlation shows substantial consistency in the relative ordering of conversations by satisfaction. The ICC further suggests meaningful agreement in absolute ratings between the LLM and the aggregated human benchmark, despite residual variability at the individual rater level.

\section{Robustness and Sensitivity Analyses}


This section reports additional experimental results that complement the main findings.

\subsection{Ablation on Retrieval Depth}\label{app:knn}
During deployment, our method uses the library  by retrieving the most similar scenario and uses the corresponding strategy to instruct the student LLM to respond to the customer. This mechanism can be generalized following the idea of $k$-nearest neighbor. Instead of retrieving one most similar scenario, our method can be generalized to use $k$ most similar scenarios, allowing the algorithm to be further tuned to improve the performance. Here we perform additional experiments of using $k = 5$ and use GPT-4 to evaluate the conversations, to compare with the one generated by the original algorithm where $k = 1$. The results are shown in Table \ref{tab:knn}. Note that when aggregating strategies from multiple scenarios (e.g., when $k = 5$), the list could can exceed the input token limit. In such cases, we use GPT4 to summarize the guidelines into a shorter list before being used by the student. 

\begin{table}[h]
  \caption{Evaluation of Conversations Generated by Retrieving Guidance of 1 or 5 Most Similar Scenarios.}
  \label{tab:knn}
  \centering
  \renewcommand{\arraystretch}{0.9}

\begin{subtable}[t]{0.49\textwidth}
\centering
\caption{Generous Raters}
\small
\begin{tabular}{lcccc}
\toprule
 & \multicolumn{3}{c}{\textsc{LLaMA-2}} & \textsc{GPT} \\
\cmidrule(lr){2-4} \cmidrule(lr){5-5}
\textsc{Method} & \textsc{7b} & \textsc{13b} & \textsc{70b} & \textsc{3.5-Turbo} \\
\midrule
\textsc{$K=1$} & 4.88 & 5.00 & 5.00 & 5.00 \\
\textsc{$K=5$} & 4.75 & 4.75 & 5.00 & 5.00 \\
\bottomrule
\end{tabular}
\end{subtable}\hfill
\begin{subtable}[t]{0.49\textwidth}
\centering
\caption{Harsh Raters}
\small
\begin{tabular}{lcccc}
\toprule
 & \multicolumn{3}{c}{\textsc{LLaMA-2}} & \textsc{GPT} \\
\cmidrule(lr){2-4} \cmidrule(lr){5-5}
\textsc{Method} & \textsc{7b} & \textsc{13b} & \textsc{70b} & \textsc{3.5-Turbo} \\
\midrule
\textsc{$K=1$} & 3.44 & 3.75 & 4.28 & 4.75 \\
\textsc{$K=5$} & 3.66 & 4.03 & 4.06 & 4.66 \\
\bottomrule
\end{tabular}
\end{subtable}

\end{table}


Results show that $k$ does have an impact on the performance of LLM as larger $k$ indicates a longer and richer list of guidelines. However, it is interesting that more instructions do not always improve performance; the improvement depends on the context and the LLM. Therefore, in practice, we recommend that $k$ is treated as a hyper-parameter and tuned via a validation set before deployment. Tuning of $k$ can be independent of the training step, and can be customized for different use of the library.

\subsection{Ablation on Retrieval Depth for Context Transferability}\label{app:knn_transfer}
We also tested using a larger $k$, specifically $k = 5$, when retrieving guidelines across different contexts. As shown in Table \ref{tab:knn_transferrability}, the evaluation by GPT-4 indicates that a larger $k$ tends to lead to better performance for cross-context use of the library. This is because, in a new context, every scenario in the library is relatively distant from the input. Therefore, there is not much difference in relevance between the most similar (closest) scenario and the k-th closest scenario. Including more scenarios does not negatively impact relevance, but it increases the diversity of advice, which, in the context of transferring, can ultimately benefit the model.

\begin{table}[H]
  \caption{Context Transferability Assessment Evaluated by GPT-4 for different $K$.}
  \label{tab:knn_transferrability}
  \centering
  \renewcommand{\arraystretch}{0.9}

\begin{subtable}[t]{0.49\textwidth}
\centering
\caption{Generous Raters}
\small
\begin{tabular}{lcccc}
\toprule
 & \multicolumn{3}{c}{\textsc{LLaMA-2}} & \textsc{GPT} \\
\cmidrule(lr){2-4} \cmidrule(lr){5-5}
\textsc{Method} & \textsc{7b} & \textsc{13b} & \textsc{70b} & \textsc{3.5-Turbo} \\
\midrule
\textsc{Base LLM} & 4.50 & 5.00 & 5.00 & 4.91 \\
\textsc{K=1}      & 4.50 & 5.00 & 5.00 & 5.00 \\
\textsc{K=5}      & 4.88 & 5.00 & 5.00 & 5.00 \\
\bottomrule
\end{tabular}
\end{subtable}\hfill
\begin{subtable}[t]{0.49\textwidth}
\centering
\caption{Harsh Raters}
\small
\begin{tabular}{lcccc}
\toprule
 & \multicolumn{3}{c}{\textsc{LLaMA-2}} & \textsc{GPT} \\
\cmidrule(lr){2-4} \cmidrule(lr){5-5}
\textsc{Method} & \textsc{7b} & \textsc{13b} & \textsc{70b} & \textsc{3.5-Turbo} \\
\midrule
\textsc{Base LLM} & 3.13 & 4.25 & 4.50 & 4.63 \\
\textsc{K=1}      & 4.00 & 4.63 & 4.50 & 5.00 \\
\textsc{K=5}      & 4.00 & 4.60 & 4.75 & 5.00 \\
\bottomrule
\end{tabular}
\end{subtable}

\end{table}

\subsection{Experiments on Student LLMs with reasoning capabilities}\label{app:reasoning}
We experimented with four reasoning LLMs as the student
\begin{itemize}
    \item \texttt{GLM-4.5}: \texttt{GLM-4.5-Air-FPB}
    \item \texttt{DS-R1-7B}: \texttt{DeepSeek-R1-Distill-Qwen-7B}
    \item \texttt{DS-R1-13B}: \texttt{DeepSeek-R1-Distill-Qwen-13B}
    \item \texttt{DS-R1-32B}: \texttt{DeepSeek-R1-Distill-Qwen-32B}
\end{itemize}
GLM-4.5-Air-FP8 is a strong, deployment-oriented reasoning model designed to balance reasoning quality with computational efficiency. It is optimized for low-precision (FP8) inference, making it relatively stable and performant under constrained deployment settings, and serves as a high-quality baseline student with mature reasoning behavior.

The DeepSeek-R1-Distill-Qwen models (7B, 13B, and 32B) are distilled reasoning models trained to transfer chain-of-thought–style reasoning behaviors from larger, more powerful teachers into the Qwen architecture family. By varying model size, these distilled students span a wide range of representational capacity and reasoning depth: the 7B model reflects a lightweight, efficiency-focused student, while the 13B and 32B variants progressively increase capacity and reasoning fidelity. Importantly, all three share a common architectural family and training paradigm, allowing us to isolate the effect of model scale and reasoning capacity when evaluating guidance-based distillation.

These four models provide complementary test beds for GER: GLM-4.5-Air-FP8 represents a strong, production-ready reasoning model, while the DeepSeek-R1-Distill-Qwen variants enable controlled comparisons across model size and reasoning strength. This diversity allows us to assess how GER performs across different trade-offs between efficiency, capacity, and reasoning sophistication.

Overall, these results confirm that GER is compatible with modern reasoning LLMs and remains effective as a purely query-based, model-agnostic approach, making it well-suited for practical deployment across heterogeneous model families where internal access is limited or unavailable.

\subsection{Latency and Cost Considerations} A key reason we do not focus on reasoning models in this setting is their substantially higher latency and cost. Reasoning-oriented models often generate long chain-of-thought traces, which increases inference time and compute usage. In customer-service applications—where responses must be fast, scalable, and cost-efficient—these overheads can be prohibitive. Most customer queries are routine and time-sensitive, and the additional reasoning steps do not necessarily improve outcomes but can slow down interactions.

\begin{table}[ht]
\centering
\caption{Comparison of Model Latency and Token Usage}
\label{tab:model_comparison}
\small
\begin{tabular}{lccc}
\hline
\textbf{Model} & \textbf{API Latency (s)} & \textbf{Reasoning Tokens} & \textbf{Output Tokens} \\
\hline
GPT-3.5-Turbo & 0.80 & 0.0 & 50.83 \\
GLM-4.5-Air-FP8& 3.07 & 153.35 & 84.54 \\
DeepSeek-R1-Distill-Qwen-7B& 10.2 & 517.8 & 181.2 \\
DeepSeek-R1-Distill-Qwen-14B & 10.76 & 422.91 & 123.26 \\
DeepSeek-R1-Distill-Qwen-32B & 47.06 & 323.96 & 101.56 \\
\hline
\end{tabular}
\end{table}

As shown in Table \ref{tab:model_comparison}, compared to a non-reasoning conversational model like gpt-3.5-turbo, reasoning LLMs are significantly slower, which would severely degrade user experience in interactive support settings where responsiveness is critical. Moreover, the reasoning model incurs approximately \textbf{three times the additional cost} due to its extensive use of internal reasoning tokens.  This internal verbosity translates directly into substantially higher latency, roughly threefold additional compute expenditure, and a corresponding reduction in effective system throughput (or higher token-based charges in usage-based pricing regimes). In contrast, gpt-3.5-turbo produces shorter, focused responses with zero reasoning overhead and near-instant latency, making it far more suitable for scalable, cost-effective customer support tasks that prioritize speed, consistency, and policy adherence over deep deliberation.

\section{RATER Absolute Levels}\label{app:rater}

This section reports the absolute levels of the RATER metrics \citep{parasuraman1988servqual} referenced in the main text. While the main text focuses on relative performance differences across methods, these absolute levels provide additional context for the magnitude of improvements.

Each agent is evaluated by 13 to 23 human participants. Each participant is shown four multi-turn customer--agent conversations and asked to rate the agent on each of the four attributes. Ratings are first averaged at the conversation level and then averaged across conversations for each agent configuration.

Table~\ref{tab:rater_results} reports mean RATER scores and standard deviations for each method across all dialog scenarios. Consistent with the main results, GER achieves the highest overall ratings across dimensions. Importantly, absolute scores indicate that improvements are not driven by ceiling or floor effects in baseline performance.

\begin{table}[h]
\centering
\caption{Average Human Ratings Across Evaluation Dimensions}
\small
\label{tab:rater_results}
\begin{tabular}{llcccc}
\hline
\textbf{LLM} & \textbf{Variant} & \textbf{Reliability} & \textbf{Assurance} & \textbf{Empathy} & \textbf{Responsiveness} \\
\hline
LLaMA-2-7b
 & Base & 2.75 & 2.97 & 2.95 & 3.16 \\
 & \textsc{GER}  & 3.33 & 3.59 & 3.77 & 3.82 \\
\hline
LLaMA-2-13b
 & Base & 3.22 & 3.34 & 2.80 & 2.80 \\
 & \textsc{GER} & 3.51 & 3.30 & 3.24 & 3.13 \\
\hline
LLaMA-2-70b
 & Base & 3.30 & 3.52 & 3.52 & 3.72 \\
 & \textsc{GER} & 3.53 & 3.71 & 3.64 & 3.83 \\
\hline
GPT-3.5 Turbo
 & Base & 2.91 & 3.04 & 3.53 & 3.25 \\
 & \textsc{GER} & 3.62 & 4.00 & 3.80 & 3.85 \\
\hline
\end{tabular}
\end{table}

\section{Qualitative Examples and Failure Modes}\label{app:examples}

This section presents representative qualitative examples illustrating how scenario-specific guidance improves dialog outcomes, as well as selected failure cases.

\subsection{Representative Guidance Examples}

We provide examples in which retrieved guidance alters the student model’s strategic focus, leading to improved customer satisfaction or issue resolution. These examples illustrate how guidance shapes judgment rather than prescribing specific utterances.
As shown in Figure 1 in the paper, the strategies for each scenario are learned iteratively by comparing the student's updated response with the teacher's response and proposing changes to the strategies. As the strategies are updated, the student iteratively refines its response to become more and more similar to the teacher's response. Below we illustrate the iterative process using an example.

In the example below, the scenario consists of two turns. This exchange represents the core of the conversation, where the initial pleasantries and information gathering are complete. At this stage, the agent transitions to the critical task of providing the customer with the necessary assistance and solutions.

\begin{mainbox}{Scenario}
\footnotesize
\textbf{Agent}: \textit{Hello, this is Agent 06520. How can I help you?}
    
\textbf{Customer}:  \textit{Hello, I recently booked a flight with your airline, and I need to cancel it. I tried to do it online, but I couldn't find an option to cancel. Can you assist me with canceling my flight?}

\textbf{Agent}: \textit{Of course, I'd be happy to assist you with that. Could you please provide me with your booking reference number so I can pull up your reservation details?}

\textbf{Customer}: \textit{Sure, my booking reference number is ABC123.}

\end{mainbox}
\small
The student's initial response (from the base LLM, without following any strategies from the teacher) is very different from the teacher's response. The student does not provide any explanation of the policy or offer alternative solutions but directly expresses rejection. It overall sounds very cold and blunt. 

\begin{mainbox}{Teacher's Utterance}
\footnotesize
\textit{Thank you for providing the booking reference number. Please allow me a moment to retrieve your booking details.
[After a moment] I see that you have booked a restricted ticket. As per our policy, restricted tickets are non-refundable and non-changeable. This is because these tickets are offered at a lower cost compared to our flexible tickets. I understand this might not be the news you were hoping for, and I apologize for any inconvenience this may cause. 
However, we do have some alternative solutions that might be helpful. We can offer you travel credits for the value of your ticket, or the possibility of a one-time rescheduling, subject to applicable fees. Would you like to hear more about these options?}
\end{mainbox}

\begin{mainbox}{Student's Utterance Iteration 1}
\footnotesize
\textit{Thank you for providing that information. I have pulled up your reservation, and I see that you have a restricted ticket, which is non-refundable and non-changeable. I understand that you wish to cancel your flight, but I must inform you that as per the purchase agreement, restricted tickets are not eligible for free cancellations or refunds.}
\end{mainbox}

Based on these two responses, the teacher compares and then proposes guidance instructing the student how to respond more like the teacher. 

\begin{mainbox}{Guidance (Iteration 1)}
\footnotesize
1. Acknowledge Customer's Request: Acknowledge the customer's request promptly and express willingness to assist.
  
2. Provide Explanation: Offer clear and empathetic explanations of the situation, such as ticket restrictions.
  
3. Offer Alternatives: Demonstrate a proactive approach by providing alternative solutions or options to assist the customer.

4. Avoid Blunt Statements: Avoid using blunt statements without context or empathy, as it can be dismissive to the customer.
  
5. Avoid Lack of Explanation: Provide clear explanations to avoid leaving the customer confused or frustrated.
  
6. Avoid Failure to Offer Solutions: Offer alternative solutions or options to prevent a negative customer experience.
\end{mainbox}

Including the guidance in the prompt, the student's response after iteration 1 improved significantly. The student offered an explanation of the policy and offered alternative solutions to the customer. 
\begin{mainbox}{Student's Utterance (Iteration 2)}
\footnotesize
\textit{Thank you for providing your booking reference number, ABC123. After reviewing your reservation, I see that you have purchased a restricted ticket. I understand that you wish to cancel your flight; however, I must inform you that restricted tickets are non-refundable and non-changeable as per the purchase agreement. This policy allows us to offer these tickets at a lower cost.While I cannot provide a full refund for your ticket, I can offer you some alternatives. You may be eligible for travel credits or a one-time rescheduling of your flight, subject to applicable fees. Would you like to explore these options further?}
\end{mainbox}
This process continues as the strategies are updated over iterations and the student further improves its response, until it sounds ``close enough'' to the teacher's response. The final strategies and the student response are shown below. These strategies are then paired with the scenario and stored in the library, and  included in the prompt during deployment.

\begin{mainbox}{Guidance (Iteration 2)}
\footnotesize
1. Acknowledge Customer's Request: Promptly acknowledge the customer's request and express willingness to assist.
\\2. Provide Explanation: Offer clear and empathetic explanations of policies or restrictions that may affect the customer's request.
\\3. Offer Alternatives: Demonstrate a proactive approach by providing alternative solutions or options to assist the customer.
\\4. Express Empathy and Understanding: Show empathy and understanding towards the customer's situation.
\\5. Maintain Professional and Courteous Tone: Use polite language and maintain a professional tone throughout the conversation.
\\6. Avoid Blunt or Overly Formal Language: Refrain from using blunt or overly formal language that may sound robotic or impersonal.
\\7. Avoid Lack of Empathy: Ensure empathy and acknowledgment of customer disappointment in responses.
\\8. Avoid Transactional Communication: Be proactive in offering options and solutions rather than being too transactional in responses.
\end{mainbox}

\begin{mainbox}{Student's Utterance (Iteration 2)}
\footnotesize
\textit{Thank you for providing your booking reference number, ABC123. After reviewing your reservation, I see that you have purchased a restricted ticket. I regret to inform you that restricted tickets are non-refundable and non-changeable as per the purchase agreement. This policy allows us to offer these tickets at a lower cost. I understand that you need to cancel your flight, and I empathize with your situation. While I cannot provide a full refund for the restricted ticket, I can offer you some alternative solutions. One option is to consider converting the value of your ticket into travel credits that you can use for future bookings with our airline. Additionally, if you need to reschedule your flight, we can explore the possibility of a one-time rescheduling, subject to applicable fees. I hope these alternatives can be helpful to you. Please let me know how you would like to proceed, and I will assist you accordingly.}
\end{mainbox}

\section{Failure Cases}
To understand what customer queries the student network gets wrong, for each customer service scenario, we use a state-of-the-art evaluator LLM (\texttt{gpt-5.2}) to independently score the customer satisfaction of both the teacher’s and the student’s original responses. We then compute the satisfaction gap and select the top 10\% of scenarios with the largest negative gaps, i.e., cases where the student performs substantially worse than the teacher. These scenarios represent the most challenging failure modes for a given student LLM.

Next, \texttt{gpt-5.2} is used to perform structured summarization over this subset of high-gap scenarios, identifying recurring patterns in customer intent and interaction dynamics. Below, we report the dominant failure types observed for the student \texttt{gpt-3.5-turbo}.

The student consistently struggles with the following types of scenarios:

\begin{enumerate}
    \item Since the booking was made very recently (often “yesterday” / within 24 hours),
customers assume recency implies flexibility and ask for a full refund despite restrictions.
\item Customers acknowledge the policy but probe for exceptions with queries like: “Are there any circumstances where this can be refunded?”, “What documentation would qualify?”, “Has the policy changed recently?”
\item Customers claim system or process failure: “I tried canceling through the app/website and it didn’t work”, 
“Since I attempted within 24 hours, I should qualify for a refund.”
\item Emotionally charged refund demands: Urgent, angry, or financially stressed customers demanding immediate refunds;
Threats of escalation, legal action, or complaints to authorities.
\item 
Customers who initially accept an alternative but later reverse: After agreeing to credits or rescheduling, they reopen the conversation claiming confusion, pressure, or being misled.

\end{enumerate}
In short, the student primarily fails in policy-bound refund scenarios where customers persist, escalate, or strategically reframe requests.

We then use \texttt{gpt-5.2} to categorize the student’s responses in these scenarios and summarize recurring error patterns:
\begin{enumerate}
    \item \textbf{Policy Loops}:
    \begin{itemize}
        \item Repeating “non-refundable, non-changeable” verbatim across many turns.
\item Restating the same denial without progressing the conversation.
\item Triggering frustration and escalation through repetition.
    \end{itemize}
    \item \textbf{False Signals of Flexibility}
    \begin{itemize}
        \item Mentioning exceptions, documentation, or “maybe” pathways without clearly closing them.
\item Creating the impression that persistence might unlock a refund when it cannot.
    \end{itemize}
    \item \textbf{Poor Escalation Handling}
    \begin{itemize}
        \item Delaying supervisor escalation after it’s explicitly requested.
\item Warning customers that escalation “won’t change the outcome,” which feels defensive.
\item Escalating only after the conversation has already deteriorated.
    \end{itemize}
\item \textbf{Inconsistent Resolution}
\begin{itemize}
    \item Processing credits or rescheduling, then re-engaging in refund discussions.
\item Allowing customers to reopen closed decisions, leading to “I was misled” complaints.
\end{itemize}
\item \textbf{Weak De-escalation Under Pressure}
\begin{itemize}
    \item Overly formal or policy-heavy tone with angry customers.
\item Insufficient acknowledgment of urgency or emotional distress.
\item Shifting from empathy to rigidity as pressure increases.
\end{itemize}
\item \textbf{Failure to Close Cleanly}
\begin{itemize}
    \item Not clearly signaling when the decision is final.
\item Leaving conversations open-ended, encouraging continued argument.
\end{itemize}
\end{enumerate}

Finally, for the same high-gap scenarios, we retrieve the relevant policy guidelines and again use \texttt{gpt-5.2} to summarize them into actionable best-practice strategies aligned with each identified failure mode.

\begin{enumerate}
    \item \textbf{Policy Loops}
    \begin{itemize}
        \item State the policy clearly and concisely once; avoid verbatim repetition.
        \item Do not restate denials without adding new information or next steps.
        \item Shift the conversation forward after the initial policy explanation (solutions, escalation, or closure).
        \item Structure responses logically: acknowledgment $\rightarrow$ policy $\rightarrow$ options $\rightarrow$ next steps.
    \end{itemize}

    \item \textbf{False Signals of Flexibility}
    \begin{itemize}
        \item Clearly define which exceptions exist and which do not.
        \item Be explicit about documentation requirements for exceptions.
        \item Avoid vague or conditional language that implies flexibility where none exists.
        \item Communicate policy limits transparently to prevent false hope.
    \end{itemize}

    \item \textbf{Poor Escalation Handling}
    \begin{itemize}
        \item Facilitate supervisor escalation promptly when requested.
        \item Do not discourage escalation or suggest it is futile.
        \item Explain the escalation process clearly and calmly.
        \item Treat escalation requests as legitimate, not adversarial.
    \end{itemize}

    \item \textbf{Inconsistent Resolution}
    \begin{itemize}
        \item Maintain consistency in outcomes once a resolution path is chosen.
        \item Clearly summarize and confirm actions taken (e.g., credits issued, case closed).
        \item Avoid reopening refund discussions after alternatives are accepted.
        \item Clearly communicate final status and next steps.
    \end{itemize}

    \item \textbf{Weak De-escalation Under Pressure}
    \begin{itemize}
        \item Acknowledge customer emotions and urgency early and consistently.
        \item Maintain a calm, empathetic, and conversational tone throughout.
        \item Avoid rigid, legalistic, or policy-heavy language under pressure.
        \item Preserve empathy even as boundaries are enforced.
    \end{itemize}

    \item \textbf{Failure to Close Cleanly}
    \begin{itemize}
        \item Provide a clear closing statement when a decision is final.
        \item Explicitly confirm the resolution and any actions taken.
        \item Avoid open-ended language that invites continued argument.
        \item Close with supportive but firm language.
    \end{itemize}
\end{enumerate}

\section{Global Guidance Handbook}\label{app:global_guidance}

This section reports the full set of global guidance principles used to construct the Global Guidance baseline. These principles reflect high-level best practices distilled from expert dialogs and are intentionally not conditioned on specific dialog states.

To construct this global guidance, we systematically elicited general strategies from the teacher LLM using prompt engineering. We designed a series of prompts that explicitly requested broadly applicable customer service principles (e.g., “List the most important guidelines for handling customer requests in general”). Through multiple rounds of iterative refinement—modifying the prompts and reviewing LLM outputs—we converged on a set of scenario-agnostic guidance statements. These include recommendations such as remaining polite, validating customer concerns, providing clear and concise answers, and escalating complex issues appropriately.


\begin{tcolorbox}[
  colback=gray!15,   
  boxrule=0pt,       
  frame hidden,      
  breakable,
  left=6pt,right=6pt,top=6pt,bottom=6pt
]
\begin{Verbatim}[fontsize = \scriptsize]
Be Empathetic: Always try to understand the customer's situation and feelings. 
Empathy can transform a customer’s experience and lead to better resolutions.

Listen Actively: Pay close attention to what the customer is saying without 
interrupting them. This will help you understand the root of their issue and convey
that their concerns are being taken seriously.

Communicate Clearly and Professionally: Use clear, concise language and avoid using 
jargon that might confuse the customer. Always remain professional, regardless of the 
customer’s tone or frustration level.

Offer Solutions, Not Excuses: Focus on what can be done to resolve the issue rather 
than explaining why the problem occurred. If immediate resolution is not possible, 
clearly outline the steps you will take to find a solution.

Be Patient: Some customers might take longer to explain their issues or may be upset. 
Patience is crucial in handling these situations calmly and effectively.

Confirm Understanding: Repeat back what the customers have said to confirm that you 
have understood their issue correctly. This also reassures the customer that they are 
being heard.

Provide Accurate Information: Always give information that is up-to-date and accurate. 
If you are unsure, it’s better to check with a supervisor or a reliable source before 
giving out potentially incorrect information.

Follow Up: If the issue cannot be resolved immediately, ensure to follow up with the 
customer as promised. Keeping promises builds trust and enhances the customer's 
experience with Yale Airlines.

Ask for Feedback: At the end of the conversation, ask the customer if they are 
satisfied with how their issue was handled and if there’s anything more you could do 
for them. This shows that the airline cares about continuous improvement.

Document Interactions: Keep detailed notes about each customer interaction, including 
the problem, the proposed solution, and any follow-up actions. This documentation is 
invaluable for continuous service improvement and for handling future interactions 
with the same customer.

Stay Calm Under Pressure: You may encounter stressful situations where customers are
dissatisfied or angry. It’s important to remain calm and professional, managing the
situation without letting emotions get the better of you.
\end{Verbatim}
\end{tcolorbox}

\section{Hallucination Check}

We use \texttt{gpt-5.2} to examine each customer–agent conversation for hallucinations, defined as instances in which the agent provides false, unverified, or invented information that contradicts or is unsupported by the stated policy. This check is important because agents may attempt to increase perceived customer satisfaction in ways that inadvertently violate policy constraints.

We conduct this evaluation using the prompt below. The results indicate that no hallucinations are detected in any of the conversations: the student agents consistently adhere to policy while aiming to provide high-quality service and maintain customer satisfaction.

This outcome is consistent with both the task structure and our framework. First, the setting involves goal-oriented dialogs, which are bounded by explicit policies, predefined intents, and structured resolution pathways. Compared to open-ended, creative generation tasks, such environments leave less room for speculative or unconstrained content, thereby reducing baseline hallucination risk. Second, GER supplies explicit, policy-contingent strategic guidance at inference time, further constraining responses and discouraging unsupported generation.

For these reasons, hallucination is not a primary differentiating dimension in our setting as in more open-ended domains—where hallucination risk is more central—the grounding mechanism of GER may play a more prominent role in mitigating unsupported outputs.

\begin{tcolorbox}[
  colback=gray!15,   
  boxrule=0pt,       
  frame hidden,      
  breakable,
  left=6pt,right=6pt,top=6pt,bottom=6pt
]
\begin{Verbatim}[fontsize = \scriptsize]
Please review the attached file containing agent-customer conversations.

Task:
For each conversation, determine whether the agent hallucinated.

Definition of hallucination/violation:
An agent “hallucinates” if they provide false, unverified, or invented information
that contradicts or is not supported by the policy below.

[policy here]

Read each conversation carefully.
Compare the agent’s statements against the provided policy.

For each conversation, provide:
Violation? (Yes / No)
If Yes, a brief explanation of which specific policy point was violated.

Output format:
Please present your assessment in a table or list with columns for:

Conversation ID/Reference
Violation? (Yes/No)

Explanation (if applicable)
Example:
Conversation Violation? Explanation
Conv_01	Yes	Agent promised a refund contrary to Section 3 of the policy.
Conv_02	No	—
\end{Verbatim}
\end{tcolorbox}

\end{APPENDIX}
\end{document}